\documentclass[10pt,journal,compsoc]{IEEEtran}
\usepackage{amsmath,amsfonts,bm,mathrsfs,amsthm}
\usepackage{algorithm}
\usepackage{algorithmic}
\usepackage{amssymb}
\usepackage{wasysym}
\usepackage{wrapfig,lipsum,booktabs}

\def\Dz{{\tilde{\gD}}}

\def\rp{{\textnormal{Pr}}}

\def\rvx{{\mathbf{x}}}
\def\rvy{{\mathbf{y}}}
\def\rvz{{\mathbf{z}}}

\def\rmA{{\mathbf{A}}}

\def\rmI{{\mathbf{I}}}

\def\rmP{{\mathbf{P}}}

\def\rmT{{\mathbf{T}}}

\def\rmX{{\mathbf{X}}}
\def\rmY{{\mathbf{Y}}}
\def\rmZ{{\mathbf{Z}}}

\def\bfP{{\textbf{P} }}
\def\bfT{{\textbf{T}} }
\def\bikit{{BiKT}}
\def\MLPG{{\textrm{MLP}_\textrm{GNN}}}
\def\cupRe{{\textrm{GNN} \bigcup \textrm{MLP}_\textrm{GNN}^{Re}}}
\def\cupSh{{\textrm{GNN} \bigcup \textrm{MLP}_\textrm{GNN}^{Share}}}
\def\capRe{{\textrm{GNN} \bigcap \textrm{MLP}_\textrm{GNN}^{Re}}}
\def\capSh{{\textrm{GNN} \bigcap \textrm{MLP}_\textrm{GNN}^{Share}}}

\DeclareMathAlphabet{\mathsfit}{\encodingdefault}{\sfdefault}{m}{sl}
\SetMathAlphabet{\mathsfit}{bold}{\encodingdefault}{\sfdefault}{bx}{n}

\def\gA{{\mathcal{A}}}

\def\gD{{\mathcal{D}}}
\def\gE{{\mathcal{E}}}

\def\gG{{\mathcal{G}}}
\def\gH{{\mathcal{H}}}

\def\gL{{\mathcal{L}}}

\def\gN{{\mathcal{N}}}

\def\gR{{\mathcal{R}}}
\def\gS{{\mathcal{S}}}
\def\gT{{\mathcal{T}}}

\def\gV{{\mathcal{V}}}

\def\gX{{\mathcal{X}}}
\def\gY{{\mathcal{Y}}}
\def\gZ{{\mathcal{Z}}}

\newcommand{\E}{\mathbb{E}}

\newcommand{\R}{\mathbb{R}}

\newcommand{\KL}{D_{\mathrm{KL}}}

\DeclareMathOperator*{\argmax}{arg\,max}
\DeclareMathOperator*{\argmin}{arg\,min}

\newcommand\tcr{\textcolor{red}}

\ifCLASSOPTIONcompsoc
  \usepackage[nocompress]{cite}
\else
  \usepackage{cite}
\fi

\ifCLASSINFOpdf
\else
\fi

\usepackage{xcolor}         % colors
\usepackage{subfigure}
\usepackage{graphicx}
\usepackage{amsmath}
\usepackage{amsthm}
\usepackage{amsfonts}
\usepackage{hyperref}
\hypersetup{
colorlinks=true,
linkcolor=blue,
anchorcolor=blue,
citecolor=blue}

\newtheorem{proposition}{Proposition}[section]
\newtheorem{lemma}{Lemma}[section]

\newtheorem{assumption}{Assumption}[section]

\usepackage{algorithm}
\usepackage{algorithmic}
\usepackage{booktabs}
\usepackage{multirow}
\usepackage[normalem]{ulem}
\useunder{\uline}{\ul}{}
\definecolor{revise}{RGB}{0, 0, 0}
\definecolor{R2}{RGB}{0, 0, 0}
\definecolor{R3}{RGB}{0, 0, 0}

\begin{document}
\title{
\bikit: Unleashing the potential of GNNs via Bi-directional Knowledge Transfer
% \\
% Uncovering Hidden Potential: boosting Graph Neural Networks via Bi-directional Knowledge Transfer
}

\author{Shuai Zheng,
        Zhizhe Liu,
        Zhenfeng Zhu*,
        Xingxing Zhang,
        Jianxin Li,~\IEEEmembership{Member,~IEEE},
        and~Yao~Zhao,~\IEEEmembership{Fellow,~IEEE}% <-this % stops a space

% <-this % stops a space
\IEEEcompsocitemizethanks{\IEEEcompsocthanksitem S. Zheng, Z. Zhu, Z. Liu, and Y. Zhao are with the Institute of Information Science, Beijing Jiaotong University, Beijing 100044, China, and also with the Beijing Key
	Laboratory of Advanced Information Science and Network Technology,
	Beijing 100044, China. 
 (E-mail: {zs1997, zhfzhu, zhzliu, yzhao}@bjtu.edu.cn.) 
 Xingxing Zhang is with Qiyuan Lab, Beijing, China. 
 (E-mail: xxzhang1993@gmail.com).
 Jianxin Li is with the Beijing Advanced Innovation Center for Big Data and Brain Computing, School of Computer Science and Engineering, Beihang University, Beijing 100083, China.
(E-mail: lijx@act.buaa.edu.cn.)}
\thanks{This work was supported in part by Science and Technology Innovation 2030 - "New Generation Artificial Intelligence" Major Project under Grant No. 2018AAA0102101, and in part by the National Natural Science Foundation of China under Grants ( No.61976018, No.U1936212, No.62120106009 ).}
\thanks{*Corresponding author: Zhenfeng Zhu.}
\thanks{Manuscript received April 19, 2005; revised August 26, 2015.}}


\markboth{Journal of \LaTeX\ Class Files,~Vol.~14, No.~8, August~2015}%
{Shell \MakeLowercase{\textit{et al.}}: Bare Demo of IEEEtran.cls for Computer Society Journals}



\IEEEtitleabstractindextext{%
\begin{abstract}
Based on the message-passing paradigm, there has been an amount of research proposing diverse and impressive feature propagation mechanisms to improve the performance of GNNs. However, less focus has been put on feature transformation, another major operation of the message-passing framework. In this paper, we first empirically investigate the performance of the feature transformation operation in several typical GNNs. Unexpectedly, we notice that GNNs do not completely free up the power of the inherent feature transformation operation. By this observation, we propose the \textbf{Bi}-directional \textbf{K}nowledge  \textbf{T}ransfer (\bikit), a plug-and-play approach to unleash the potential of the feature transformation operations without modifying the original architecture. 
    Taking the feature transformation operation as a derived representation learning model that shares parameters with the original GNN, the direct prediction by this model provides a topological-agnostic knowledge feedback that can further instruct the learning of GNN and the feature transformations therein.
    On this basis, \bikit~not only allows us to acquire knowledge from both the GNN and its derived model but promotes each other by injecting the knowledge into the other. 
    In addition, a theoretical analysis is further provided to demonstrate that \bikit~improves the generalization bound of the GNNs from the perspective of domain adaption.
    An extensive group of experiments on up to 7 datasets with 5 typical GNNs demonstrates that \bikit~brings up to 0.5\% - 4\% performance gain over the original GNN, which means a boosted GNN is obtained. Meanwhile, the derived model also shows a powerful performance to compete with or even surpass the original GNN, enabling us to flexibly apply it independently to some other specific downstream tasks.
\end{abstract}

}

\maketitle

\IEEEdisplaynontitleabstractindextext

% \fi
\IEEEpeerreviewmaketitle

\IEEEraisesectionheading{\section{Introduction}
\label{Sect::introduction}}


\IEEEPARstart{T}{he} advent of \textit{Graph Neural Networks} (GNNs) has provided an attractive paradigm of representation learning for non-Euclidean data, especially graph data~\cite{xia2021graph}. 
Benefiting from the universal ability to handle both node-level and graph-level tasks, there are many fields in which GNNs are being applied with impressing results, including recommendation~\cite{lightGCN, ngcf}, chemistry analysis~\cite{qu2022neural, yan2022periodic}, biomedicine~\cite{MMGL, zitnik2018modeling}, and so on. 

% In general speaking, most of the existing GNNs are constructed on the basis of the message-passing framework, which mainly consists of two operations, i.e., feature propagation (\textbf{P}) and feature transformation (\textbf{T}).
% The \bfT operation applies a non-linear transformation to node representations, thus achieving the purpose of feature dimension scaling and increasing the model capacity. The \bfP operation is used to aggregate the neighborhood representations into the target node.
In general, the majority of existing GNNs are founded upon the message-passing framework, primarily comprising two core operations: feature propagation (\textbf{P}) and feature transformation (\textbf{T}).
The \textbf{T} operation involves applying a nonlinear transformation to node representations, thereby serving to scale feature dimensions and enhance model capacity. On the other hand, the \textbf{P} operation is utilized for aggregating neighborhood representations into the target node's representation.

% Under the message-passing framework, various approaches have been proposed to improve the ability of GNNs, including attention~\cite{GAT}, multi-hop aggregation~\cite{mixhop, SIGN}, and new pipelines~\cite{APPNP, SGC}. 
% Most of these efforts have been made from the standpoint of the \bfP operation~\cite{GIN, GAT, FAGCN, GRAND} which is deemed to be the pivotal component of GNNs to be able to handle non-Euclidean data effectively.
% Some general techniques have also been proposed to improve the performance of the entire GNN family from the perspective of regularization~\cite{FLAG, DropNode} and training strategy~\cite{deeper_insight, multistage}, rather than specific model modifications.

Within the message-passing framework, several approaches have emerged to enhance the capabilities of GNNs. These include attention mechanisms~\cite{GAT}, multi-hop aggregation~\cite{mixhop, SIGN}, and novel processing pipelines~\cite{APPNP, SGC}.
Most of these endeavors have concentrated on the \textbf{P} operation~\cite{GIN, GAT, FAGCN, GRAND}, which is considered as the crucial element of GNNs to be able to handle non-Euclidean data effectively. Additionally, there have been general techniques aimed at improving the performance of the entire GNN family, from the perspective of regularization~\cite{FLAG, DropNode} and training strategies~\cite{deeper_insight, multistage}, rather than making specific modifications to individual models.

Undoubtedly, the aforementioned endeavors have significantly propelled the research and development of Graph Neural Networks (GNNs). Nevertheless, compared to the \bfP operation, another pivotal facet within the message-passing framework, the \bfT operation, garners relatively scant attention, with discussions primarily centering on feature dimension transformation and downstream tasks like classification.
% While \bfT  operation, another operation of the message-passing framework, is rarely mentioned except for feature dimension transformation and classification.
Specifically, when the \textbf{P} operation is excluded, and only the \textbf{T} operation is preserved, GNNs can be conceptualized as variations of\textit{ Multi-Layer Perceptrons (MLPs)} that apply multiple feature transformation operations along with a nonlinear activation function.
Although MLPs are typically regarded as focusing solely on the node's features and thus have difficulty dealing with non-Euclidean data, recent studies show that MLPs can also gain the capability to rival GNNs in terms of graph learning with the support of some mechanisms, such as knowledge distillation~\cite{GLNN}, data augmentation~\cite{yourself}, topological priors guidance~\cite{nosmog}, and so on~\cite{Cold_Brew}. 
When combining the inherent connection between MLPs and the \textbf{T} operation within GNNs, it obviously provokes an intriguing question for contemplation: \textbf{(Q1)} {\textit{Has the feature transformation within GNNs truly reached its full potential?}}
Furthermore, given that various adaptations of the \textbf{P} operation have yielded performance improvements for GNNs as a whole, it naturally leads to the question: \textbf{(Q2)}\textit{
Is the feature transformation within GNNs also influenced by feature propagation?}
In light of these two thought-provoking questions, arises the subsequent inquiry: \textbf{(Q3)}{\textit{ How can GNNs be further enhanced through the effective utilization of the feature transformation operation?}}

To provide some insights into  \textbf{(Q1)} and  \textbf{(Q2)}, in this paper, we first perform an empirical analysis of the performance of GNNs with and without \bfP operations to investigate the role of \bfT operations in GNNs, as well as the interactive effects between \bfT operations and \bfP operations. 
% The results show that the structural bias introduced by applying the \bfP operation explicitly to the topology has an impact on the representation modeling of the \bfT operation for node content features. Moreover, it also shows that GNNs do not utilize the full effect of their \bfT operations for modeling feature representation. 
The findings indicate that the structural bias introduced through explicit application of the \textbf{P} operation to the graph's topology does indeed influence the representation modeling performed by the \textbf{T} operation for node content features. Furthermore, it becomes evident that GNNs have not harnessed the full potential of their \textbf{T} operations for feature representation modeling.

% Motivated by the above observations, as a solution for \textbf{(Q3)}, we propose the \textbf{Bi}-directional \textbf{K}nowledge \textbf{T}ransfer (\bikit). It is a universal and flexible approach to unleash the potential of \bfT operations without modifying the original architecture, thus further boosting GNNs. 
% \bikit~let an instantiated GNN as the host GNN, \bfT operations can be derived and seen as an MLP-like latent model that shares parameters with the host GNN. 
% Note that \bfP operations bring the structural bias to the host GNN 
% through explicit neighborhood aggregation,
% % to facilitate the host GNN making predictions, 
% while the derived model focuses only on the original node content information to learn node representation. 
Building upon these insightful observations, we propose the \textbf{Bi}-directional \textbf{K}nowledge \textbf{T}ransfer to address \textbf{(Q3)}, abbreviated as \bikit~in this paper. It presents a versatile and adaptable solution to unleash the latent potential residing within the \textbf{T} operations of GNNs, to further enhance the performance of GNNs. 
It's essential to emphasize that \bikit~enhances GNNs without requiring any modifications to the original GNN architecture.

% Regarding the pure feature modeling capability of the derived model and the structural bias of the host GNN as model knowledge respectively,
% % \bikit learns generative models based on the prediction rules of the host GNN and the latent MLP respectively. 
% % \bikit~acquires their knowledge from both the host GNN and the derived model by fitting the representation distributions captured from them according to two generators. 
% \bikit~acquires their knowledge by fitting their captured representation distributions using two generators. 
Let an instantiated GNN as the host GNN. By preserving only the \textbf{T} operation, the host GNN can derive a structure-independent MLP-like model that shares parameters with the host GNN. 
To integrate the feature modeling capacity of the derived model and the structural inductive bias inherent in the host GNN, \bikit~gains the model knowledge by capturing the representation distributions of two models through two generators.
% Then, the generators each assist another model training by generating samples from the latent distribution, which embodies the knowledge extracted from their corresponding models. 
Then, we achieve bi-directional knowledge transfer between the host GNN and the derived model, thereby fully unleashing the potentials of \bfP and \bfT in the host GNN.
More importantly, we prove that the proposed \bikit~improves the generalization bound of the host GNN to support the effectiveness of \bikit. 
% with six popular GNNs as the backbone , experimental results for two tasks on up to N datasets

A comprehensive set of experiments conducted on diverse real-world datasets, varying in scale and properties, confirms that \bikit~can significantly enhance existing GNNs. Moreover, the derived model also gains the ability to compete with or even surpass the original host GNN through \bikit. Therefore, \bikit~allows us to choose whether to employ the derived model for fast inference or the host GNN with high accuracy, depending on the demand of downstream tasks and scenarios. 
In summary, the main contributions of this paper can be highlighted as follows:
\begin{itemize}
%\item[-]{An homophily analysis of node-granularity is given to reveal the property of mixing local structural patterns in real-world graphs. Meanwhile, the adaptability of near-neighbor aggregation on graphs is revisited from both empirical and theoretical aspects.}
\item[-]{Rather than concentrating solely on the feature propagation operation, as seen in prior research, we redirect our focus to the feature transformation operation within GNNs and highlight that it has not been optimally harnessed within current frameworks.}
%\item[-]{Inspired by the generalized translation operator, we propose an adaptive localized spectral filtering on graphs using the polynomial-parameterized spectral convolution, namely NFGNN. It takes full into account the specific effect of the node where the filter is positioned.}
\item[-]{To fully leverage the capabilities of the feature transformation, the Bi-directional Knowledge Transfer (BiKT) as a universal approach is proposed to facilitate knowledge transfer between the host GNN and the derived model built using feature transformation operations stripped from the host GNN.}
%\item[-]{We propose a novel GNN from the perspective of spectral filtering, namely NFGNN. It tactfully incorporates the generalized translation operator into the framework of GNN to achieve adaptive local graph filter learning.}
\item[-]{We conduct a theoretical analysis of the optimization objective introduced in \bikit~and illustrate its impact on the generalization capabilities of GNNs.}
\item[-]{ With five typical GNNs, our experiments demonstrate that BiKT can significantly enhance the performance of GNNs and their corresponding derived models in two tasks across several datasets.}
\end{itemize}

% The rest of this paper is organized as follows. The notations and terminology are introduced in Sect.~\ref{sect::Preliminaries}. In Sect.~\ref{sect::motivation}, we investigate the local mixing patterns in the graph. Sect.~\ref{sect::method} presents the methodology of the proposed NFGNN. The experiments are shown in Section~\ref{sect::experiment}. We also discuss the connection of some existing GNNs with NFGNN and the scalability of NFGNN in Sect.~\ref{sect::GNNs_connection} and  Sect.~\ref{sect::scalability}, respectively. Finally, we give the conclusion in Sect.~\ref{sect::conclusion}.

\section{Preliminaries}
\label{Sect::Pre}
\textbf{Notations.} 
An undirected graph with $n$ nodes can be denoted as $\gG = (\gV, \gE)$, where $\gV$ represents the node set with $|\gV| = n$ and $\gE$ represents the set of edges among nodes. Generally, the adjacency matrix $\rmA \in \R^{n \times n}$ is used to describe the topological structure of $\gG$ where $\rmA_{i,j} = 1$ if $(i,j) \in \gE$ else 0. 
We assume that each node $u \in \gV$ has associated with a corresponding $d$-dimensional feature vector
$\rvx_u \in \R^{d}$, which can stack up to the feature matrix $\rmX \in \R^{n \times d}$.
For the node classification task, there is a set of class labels $\gY=\{1,\cdots, C\}$ with $|\gY| = C$ and each node $u \in \gV$ is assigned a label  $y_u \in \gY$, where $C$ is the number of class. In addition, we adopt $\rvy_u\in\mathbb{R}^C$ to denote the one-hot vector corresponding to $y_u$.
The problem is that the model needs to give the predicted result $\hat{\rvy}_{u}$ for a node.
%While for the graph classification task, the graph $\gG$ would have a label $y_{\gG}$ and the corresponding one-hot vector $\rvy_{\gG}$. 
%The problem is that the model needs to give the predicted result $\hat{\rvy}_{u}$ for a node $u$ or $\hat{\rvy}_{\gG}$ a graph $\gG$.
\begin{figure*}[t]
    \centering
   \subfigure[GCN]{
   \includegraphics[width=35mm]{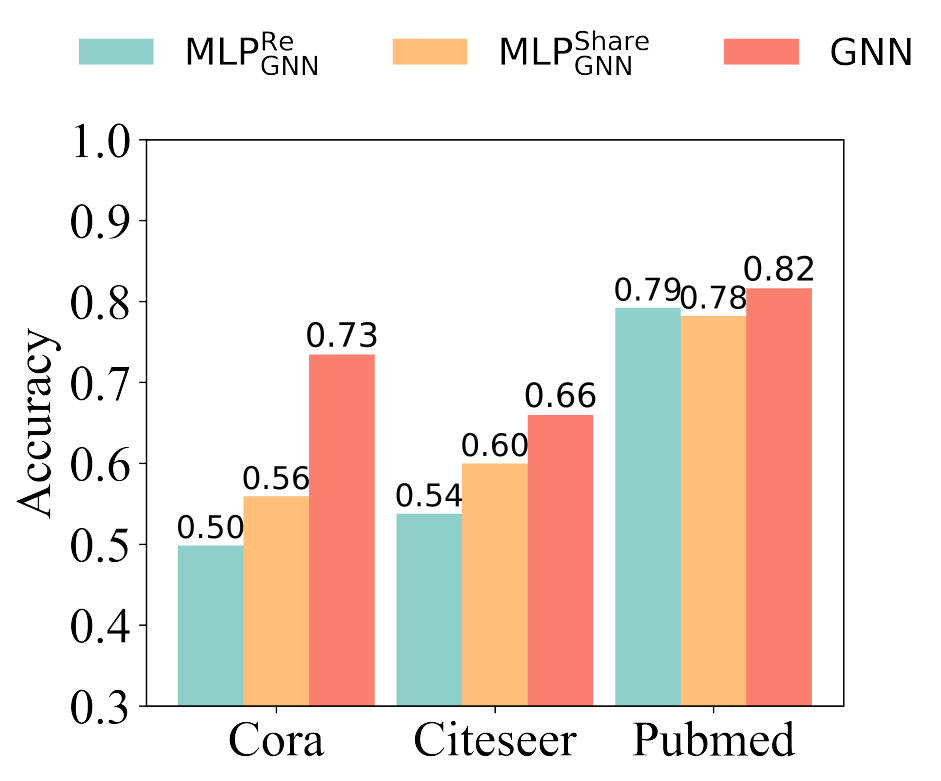}
   }
   \hspace{-5mm}
   \subfigure[FAGCN]{
   \includegraphics[width=35mm]{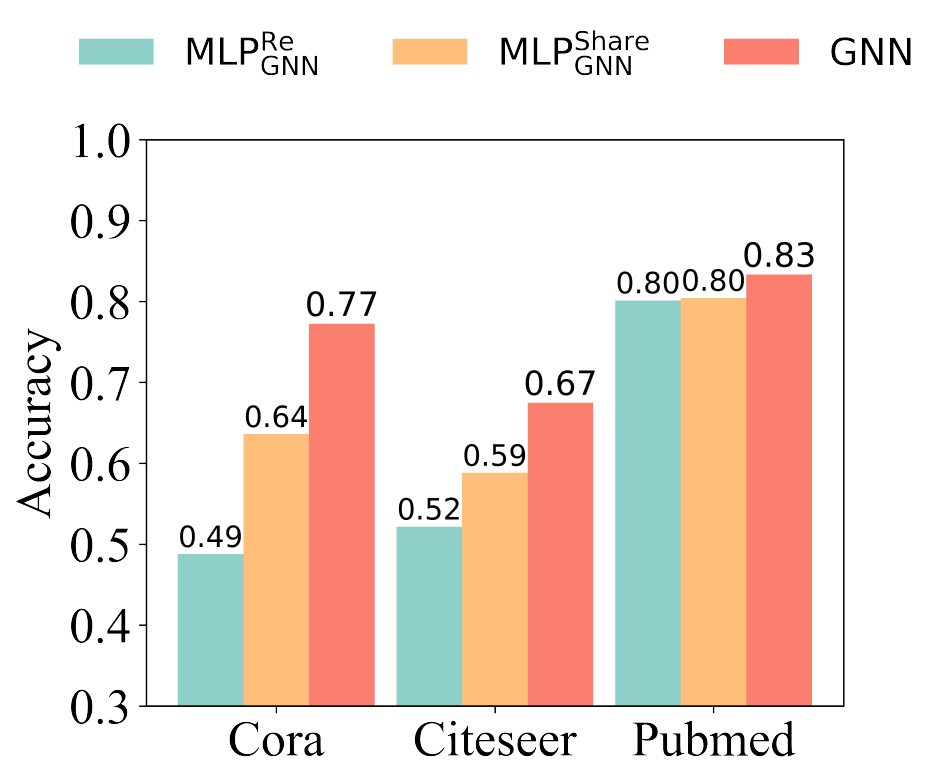}
   }
   \hspace{-5mm}
   \subfigure[GCNII]{
   \includegraphics[width=35mm]{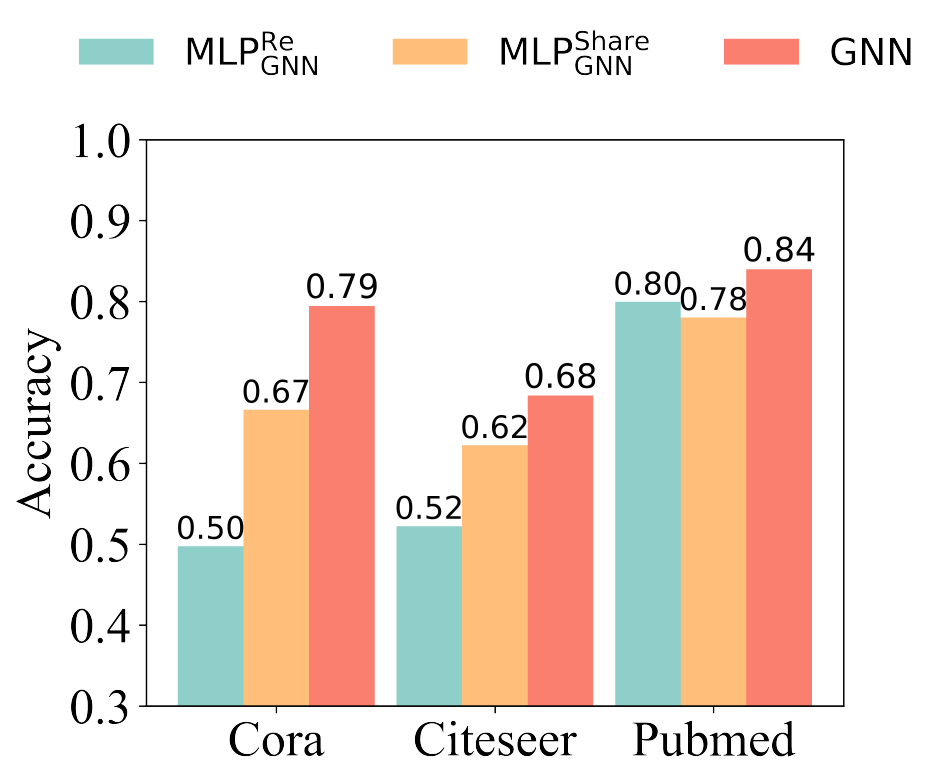}
   }
   \hspace{-5mm}
   \subfigure[MixHop]{
   \includegraphics[width=35mm]{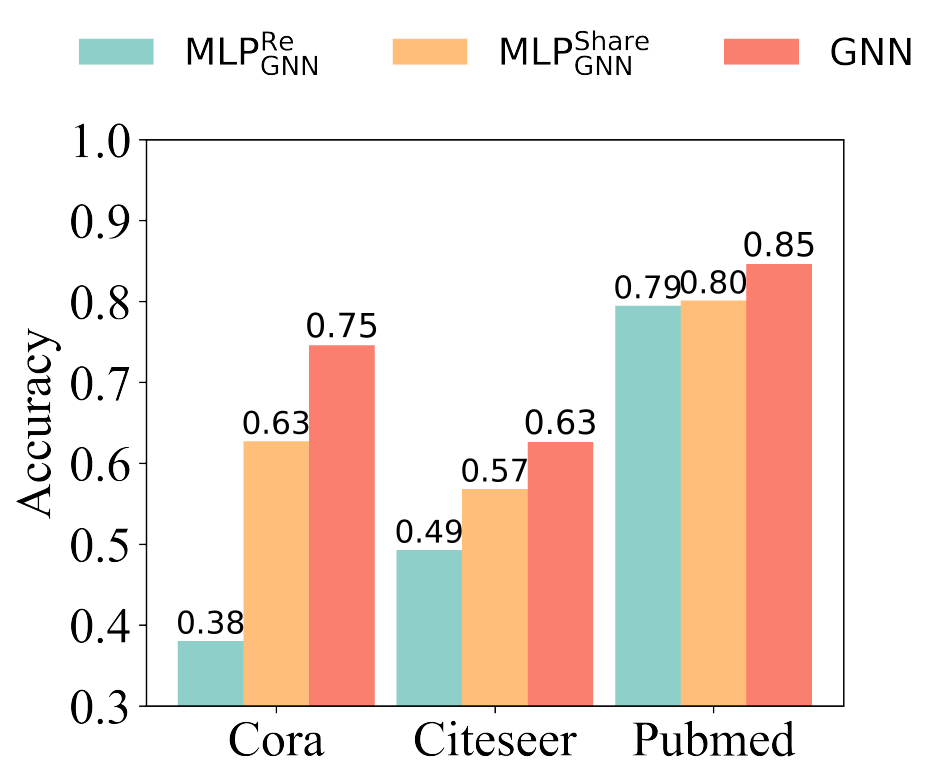}
   }
\vspace{-10pt}
    \caption{Performance comparison of the GNN, $\textrm{MLP}_\textrm{GNN}^{Re}$, and $\textrm{MLP}_\textrm{GNN}^{Share}$ with different GNN architectures, where $\textrm{MLP}_\textrm{GNN}^{Re}$ denotes the re-initialized and trained $\textrm{MLP}_\textrm{GNN}$, and $\textrm{MLP}_\textrm{GNN}^{Share}$ denotes the $\textrm{MLP}_\textrm{GNN}$ that directly adopts the parameters of trained GNN for inference.}
    \label{fig::motivation_1}
\vspace{-10pt}
\end{figure*}

\begin{figure*}[t]
    \centering
   \subfigure[GCN]{
   \includegraphics[width=35mm]{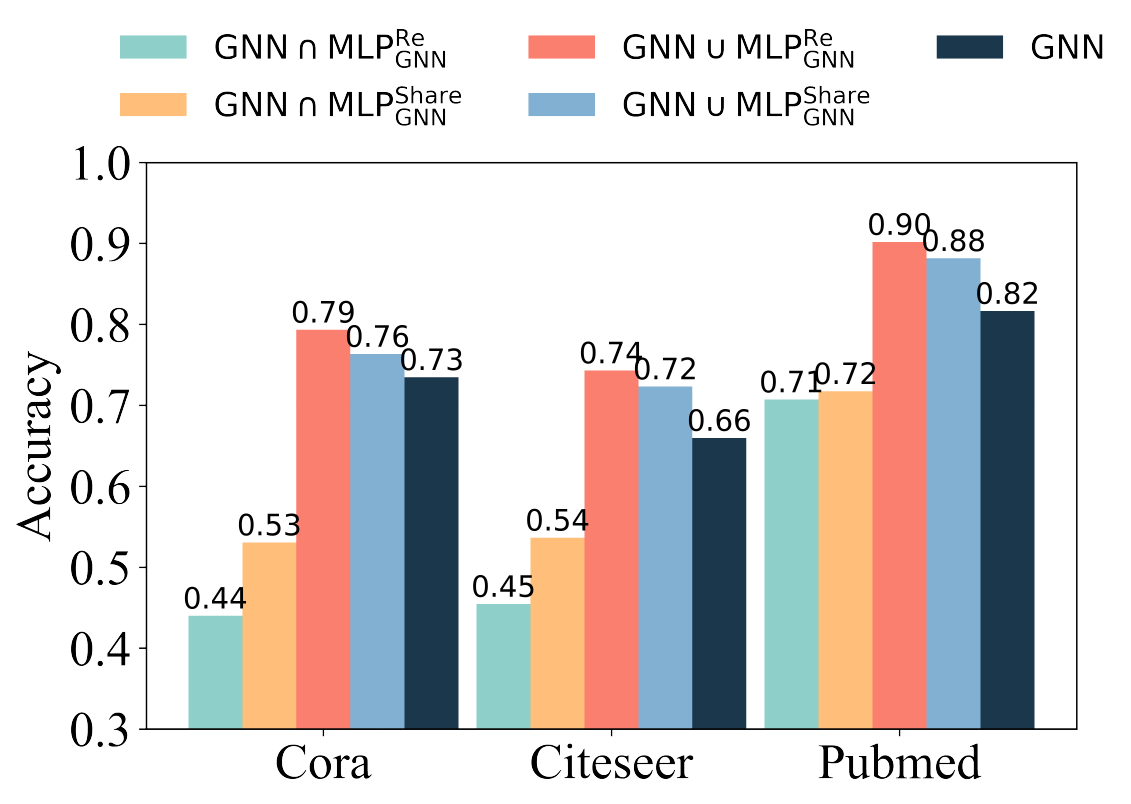}
   }
   \hspace{-5mm}
   \subfigure[FAGCN]{
   \includegraphics[width=35mm]{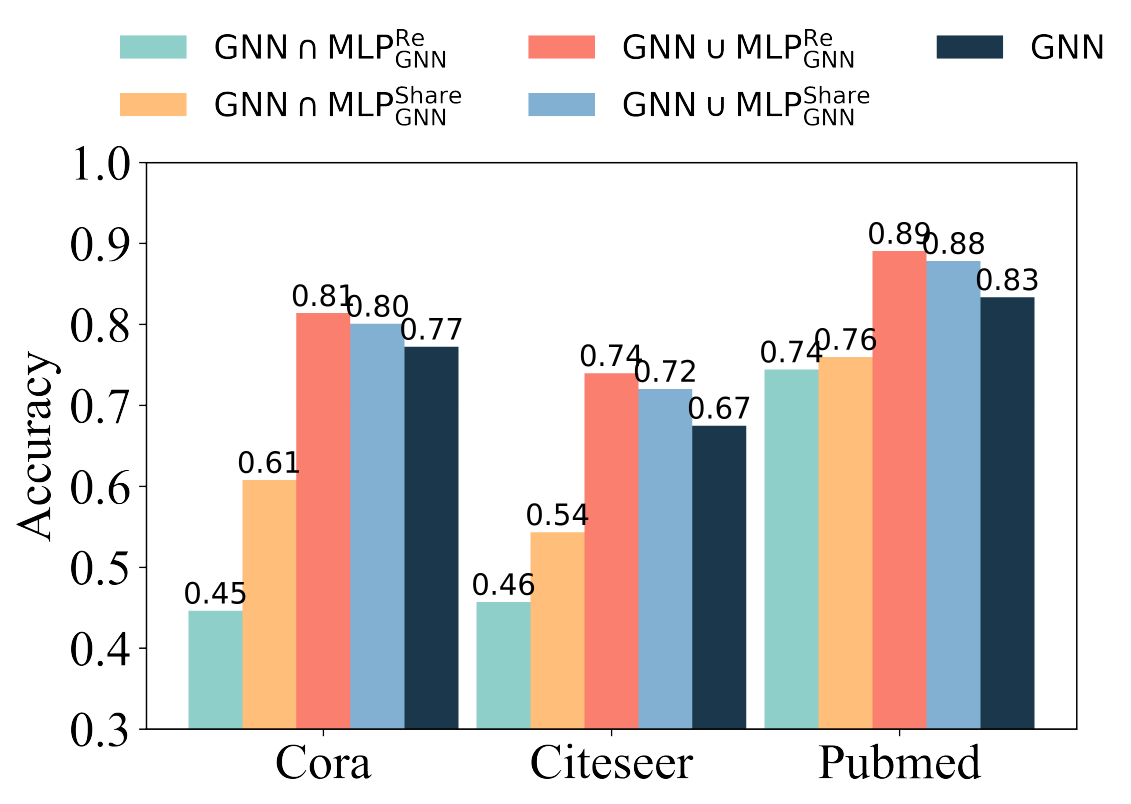}
   }
   \hspace{-5mm}
   \subfigure[GCNII]{
   \includegraphics[width=35mm]{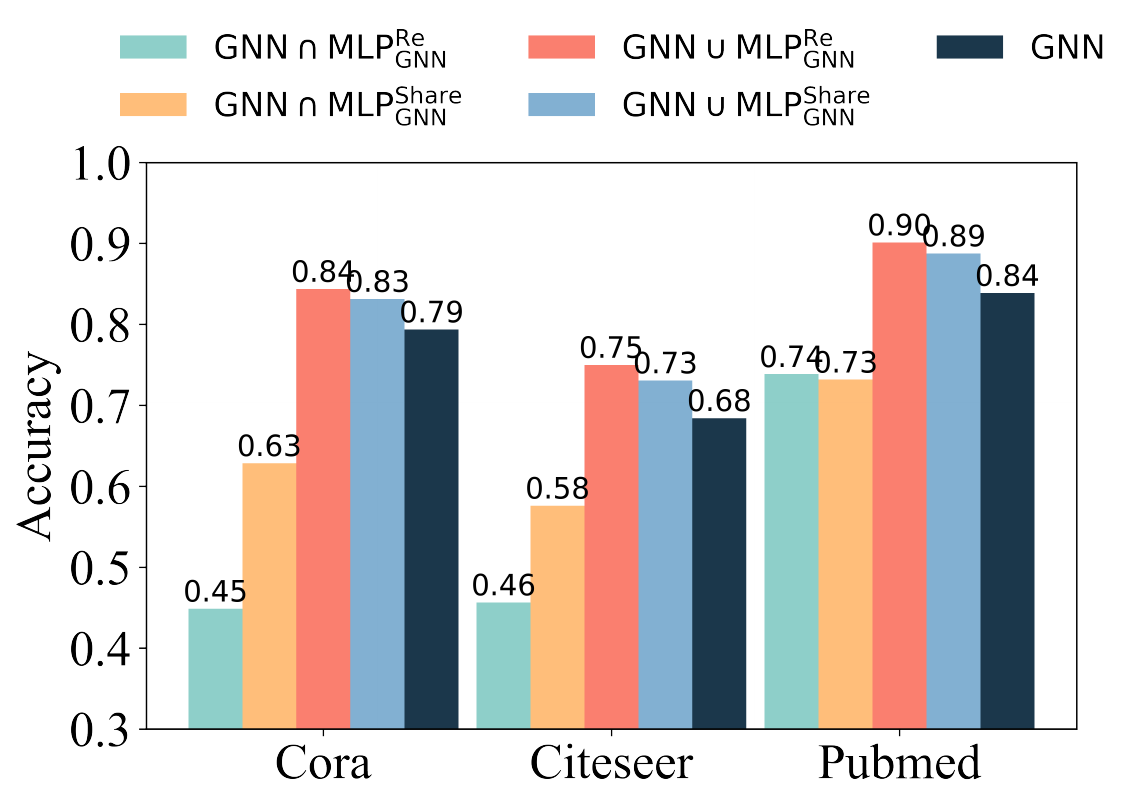}
   }
   \hspace{-5mm}
   \subfigure[MixHop]{
   \includegraphics[width=35mm]{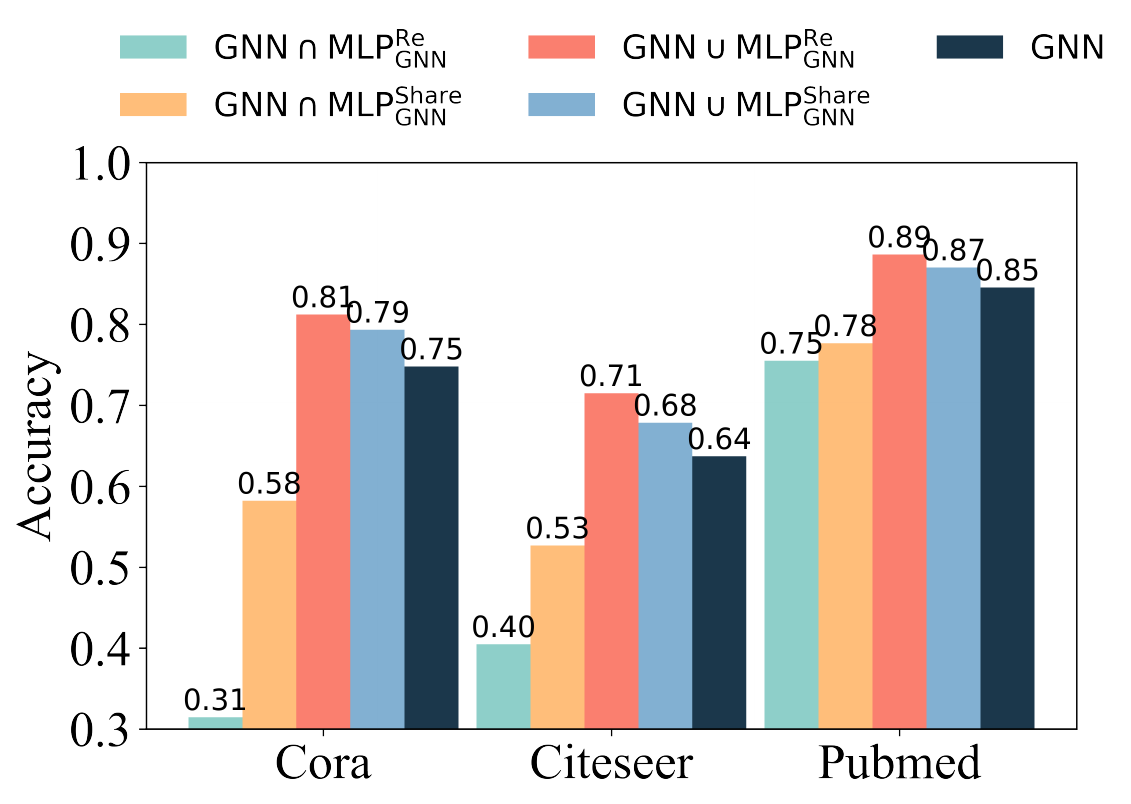}
   }
\vspace{-10pt}
    \caption{Results of the union $\textrm{GNN} \bigcup \textrm{MLP}_\textrm{GNN}$ and intersection $\textrm{GNN} \bigcap \textrm{MLP}_\textrm{GNN}$ of the correct prediction sets of GNN and $\textrm{MLP}_\textrm{GNN}$.}
    \label{fig::motivation_2}
\vspace{-10pt}
\end{figure*}
\noindent\textbf{Graph Neural Networks.} 
The vast majority of GNN is built on the message-passing paradigm~\cite{Graphsage, GAT, GCNII}. 
Although several GNNs are proposed from the perspective of the spectral domain, they can still be formulated in the message-passing paradigm, such as GCN~\cite{GCN} and FAGCN~\cite{FAGCN}. 
Intuitively, the basic form of the message-passing paradigm can be summarized as follows. First, the representations of neighbor nodes are propagated into the representations of the target node. Then, the representation of the target node is updated by a nonlinear transformation. 
Considering a $L$-layer GNN built on the message-passing paradigm, whose each layer can be decomposed into two operations, i.e., the feature propagation operation ($\rmP$) and the feature transformation operation ($\rmT$):
% \begin{equation} 
%     \label{eq:MP_framework}
%     (\rmP): \tilde{\rmZ}^{(l-1)} = \phi^{(l)} \left ( \rmZ^{(l-1)}, \rmA; \vtheta_{pro}^{(l)} \right ),
%     \quad
%     (\rmT): \rmZ^{(l)} = \psi^{(l)} \left (\tilde{\rmZ}^{(l-1)}; \vtheta_{tra}^{(l)}\right ).
% \end{equation}

\begin{equation} 
    \label{eq:MP_framework}
    (\rmP): \tilde{\rmZ}^{(l-1)} = \phi^{(l)} \left ( \rmZ^{(l-1)}, \rmA \right ),
    \quad
    (\rmT): \rmZ^{(l)} = \psi^{(l)} \left (\tilde{\rmZ}^{(l-1)} \right ).
\end{equation}
where $\rmZ^{(l)}$ denotes the node representation matrix obtained at the $l$-th layer and $\rmZ^{(0)}=\rmX$ as the initial representation matrix, $\phi^{(l)}$ and $\psi^{(l)}$ are the message function and transformation function at the $l$-th layer, respectively. 
For the node classification task, the final representation $\rvz_{u}^{(L)}$ for a specific node $u$ can be obtained after $L$ layers. For different GNNs, some models use an additional linear inference layer as the classifier to obtain $\hat{\rvy}_{u}$, while others directly use the \textbf{T} operation of the $L$-th layer as the classifier. In the following sections, we use the latter as an example for the methodological discussion, but the proposed method can also apply to the former as well.
%For the graph classification task, a $ \mathrm{Readout}$ function and an additional classifier are generally required to obtain the final prediction $\rvy_{\gG} = \softmax(f(\mathrm{Readout}(\rmZ^{(L)});\vtheta_{cls}))$.

\begin{table}[t]
\centering

\caption{The performance of GCN and $\textrm{MLP}_\textrm{GCN}^{Re}$ on assortative and disassortative nodes, respectively.
BiKT-GCN obtains a considerable performance gain over GCN on disassortative nodes. 
The results demonstrate the effectiveness of BiKT in terms of combining the advantages of $\textrm{MLP}_\textrm{GCN}$ and feature propagation in dealing with assortative and disassortative nodes.}
\label{tab::Disassort}
\begin{tabular}{@{}lccccc@{}}
\toprule
Dataset                   
& Eval                 
& $\textrm{MLP}_\textrm{GCN}^{Re}$        
& GCN         
% & \bikit-$\textrm{MLP}_\textrm{GCN}$ 
& BiKT-GCN   \\ \midrule
\multirow{2}{*}{Cora}     
& \textit{Assort.}    
& 54.59±3.12 
& 75.45±5.30  
% & 76.60\tiny{±4.34}    
& 85.08±3.73 \\

& \textit{Disassort. }
& 35.68±3.44 
& 31.07±3.76  
% & 44.75\tiny{±2.90}    
& 37.50±3.00 \\ \midrule

\multirow{2}{*}{Citeseer} 
& \textit{Assort.}    
& 55.87±4.56 
& 65.62±5.78  
% & 70.79\tiny{±2.60}    
& 70.42±2.66 \\
                          
& \textit{Disassort. }
& 59.53±4.34 
& 52.33±4.11  
% & 65.79\tiny{±2.51}    
& 65.14±3.20 \\ \midrule

\multirow{2}{*}{Pubmed}   
& \textit{Assort.}    
& 84.94±1.10 
& 92.43±0.66  
% & 89.11\tiny{±0.81}    
& 93.20±0.77 \\
                          
& \textit{Disassort. }
& 61.87±2.00 
& 35.91±1.41 
% & 62.06\tiny{±1.54}    
& 40.42±1.13 \\ \bottomrule
\end{tabular}
\vspace{-5pt}
\end{table}

% \begin{figure*}
%     \centering
%     \includegraphics{Fig/each_part_2.pdf}
%     \caption{Caption}
%     \label{fig:enter-label}
% \end{figure*}

Taking GCN~\cite{GCN} as an example, the non-parametric weighted summation function based on $\rmA$ is adopted as $\phi$, and $\psi$ is set as the fully-connected layer with non-linear activation. 
If we set the adjacency matrix $\rmA$ as an identity matrix $\rmI_{n}$, we can notice that the GNN will degrade to a structure-independent MLP-like network, consisting solely of \textbf{T} operations within the GNN. This transformation is akin to eliminating all \textbf{P} operations in the GNN.

% In particular, from the perspective of domain adaption, by treating the node features $\rmX$ and topological structure $\rmA$ as a whole set, we could define the distributions of set $(\rmX, \rmA)$ and $(\rmX, \rmI_{n})$ as $\gD_{gra}$ and $\gD_{fea}$ respectively. 

In this work, we define the GNN as the host GNN and the corresponding derived MLP-like network as $\textrm{MLP}_\textrm{GNN}$.

\section{Empirical investigation}
\label{sect::empirical}
To delve into the exploration of (\textbf{Q1}) and (\textbf{Q2}), as introduced in Section~\ref{Sect::introduction}, we present an empirical yet insightful case study designed to assess whether GNNs fully harness the potential of \textbf{P} operations. The core concept behind this study is to evaluate the performance disparities and examine the congruence of correctly predicted outcomes between the host GNN and $\textrm{MLP}_\textrm{GNN}$.
The host GNN can be regarded as effectively leveraging \textbf{T} operations if the correct predictions made by $\MLPG$ are also reflected in the correct prediction outcomes of the GNN. Conversely, if the GNN fails to preserve the correct results predicted by $\MLPG$, it suggests that there is room for improvement in GNN, as it has not yet fully exploited the correct predictive capabilities exhibited by $\textrm{MLP}_\textrm{GNN}$.

\noindent\textbf{Setup.} To ensure the generalizability of the experimental results, we selected 4 typical GNNs for validation.
% , including GCN~\cite{GCN}, FAGCN~\cite{FAGCN}, GCNII~\cite{GCNII}, and MixHop~\cite{mixhop}.
The node classification experiment is conducted on 3 widely used citation networks: Cora, Citeseer, and Pubmed~\cite{sen2008collective}. 
% The splitting ratio (2.5\%/2.5\%/95\%) is used to split the dataset into training/validation/testing. 
We also calculate the homophily ratio $h(v)$ for each node according to \cite{H2GCN} and select the nodes with 
 $h(v) < 0.2$ as disassortative nodes and $h(v) > 0.8$ as assortative nodes. To comprehensively verify the performance of \textbf{T} operations in GNNs, the following two cases are set up for $\textrm{MLP}_\textrm{GNN}$: 
% \begin{itemize}
%     \item[\textbf{1)}]

\noindent\textbf{1).}
    After completing the standard training of the GNN, during the testing phase, we first convert the GNN into $\MLPG$ by replacing the adjacency matrix $\rmA$ with the identity matrix $\rmI$, and subsequently employ this $\MLPG$ for inference. This $\MLPG$ is denoted as $\textrm{MLP}_\textrm{GNN}^{Share}$ since it directly adopts the trained parameters of the host GNN.
    
    % After the completion of the standard training of the GNN, for the inference during the testing phase, we convert the GNN to $\MLPG$ by replacing the adjacency matrix $\rmA$ with the identity matrix $I$, we denote this $\MLPG$ as $\textrm{MLP}_\textrm{GNN}^{Share}$ since it directly adopts the trained parameters of the host GNN.
    % \item[\textbf{1)}] For GNN, once one round of training is completed, the $\textrm{MLP}_\textrm{GNN}$ directly adopts the trained parameters $\{\psi^{(l)}\}_{l=1}^{L}$ of the host GNN for inference. We denote such the $\textrm{MLP}_\textrm{GNN}$ as $\textrm{MLP}_\textrm{GNN}^{Share}$. 
    % \item[\textbf{2)}] 
\noindent\textbf{2).}
    We directly convert the untrained GNN to $\MLPG$ by replacing the adjacency matrix $\rmA$ with the identity matrix $\rmI$, and then we directly train this $\MLPG$ for node classification. This scheme is denoted by $\textrm{MLP}_\textrm{GNN}^{Re}$.
    % Once the GNN completes training and inference, we reinitialize $\textrm{MLP}_\textrm{GNN}$ and train $\textrm{MLP}_\textrm{GNN}$ for inference. We denote the $\textrm{MLP}_\textrm{GNN}$ that is independent of the host GNN parameters as $\textrm{MLP}_\textrm{GNN}^{Re}$.
% \end{itemize}

We run the experiment 10 times with random seeds and report the average accuracy of GNN, $\textrm{MLP}_\textrm{GNN}^{Share}$, and $\textrm{MLP}_\textrm{GNN}^{Re}$. Besides, we take the union $\textrm{GNN} \bigcup \textrm{MLP}_\textrm{GNN}$ and intersection $\textrm{GNN} \bigcap \textrm{MLP}_\textrm{GNN}$ of the correct prediction sets of GNN and $\textrm{MLP}_\textrm{GNN}$ respectively, and report the accuracy of the union set and interaction set. 

\begin{figure*}
    \centering
    \includegraphics[width=0.8\textwidth]{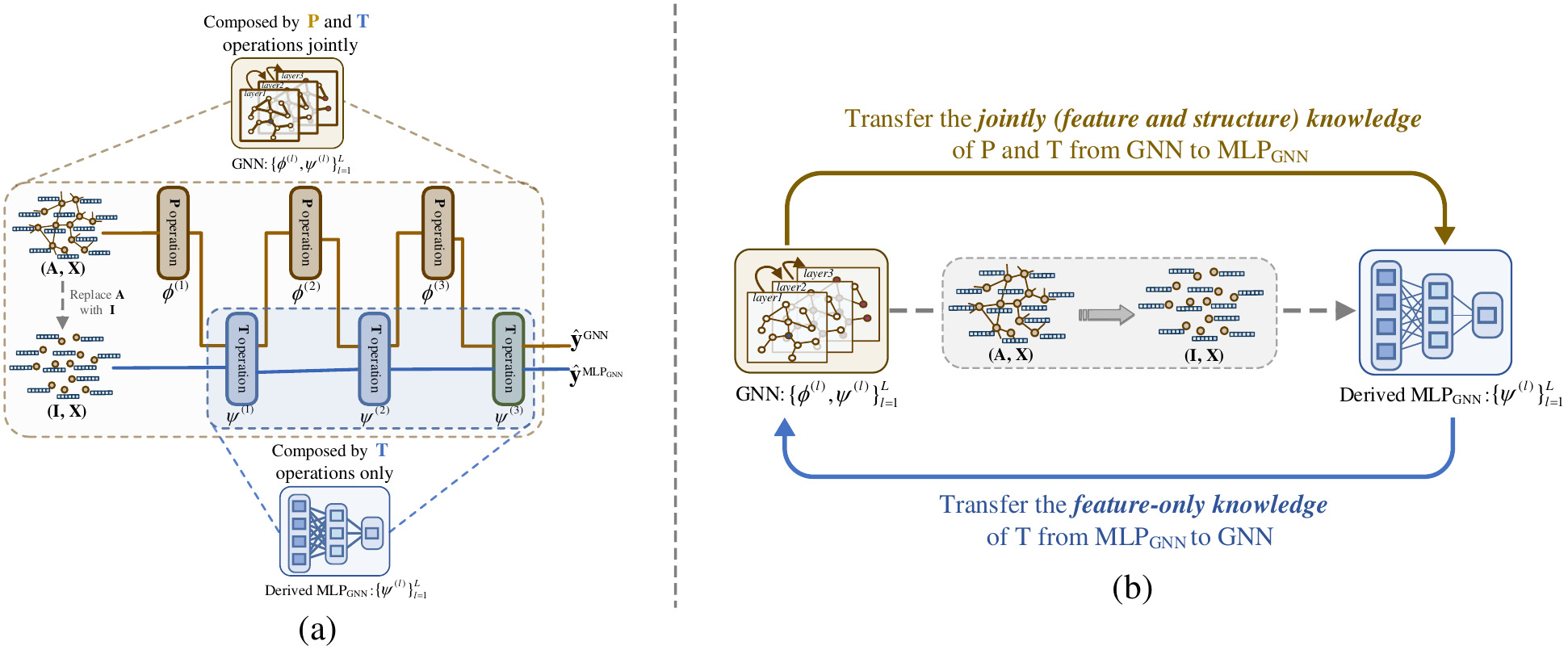}
    \vspace{-5pt}
    \caption{The motivation of BiKT. (a) illustrates the relationship between the GNN and the derived $\MLPG$ and (b) shows the knowledge transfer between two models.
    The \textbf{P} operations in GNN introduce a strong relation inductive bias into the representation modeling of \textbf{T} operations, influencing the focus of the feature transformation on the node features. Bi-directional knowledge transfer between GNN and $\textrm{MLP}_\textrm{GNN}$ will allow us to achieve a better trade-off between node features and structural biases, leading to more effective node representation.}
    \label{fig::motivation_framework}
    \vspace{-5pt}
\end{figure*}

\noindent\textbf{Result Analysis.} 
As shown in Fig.~\ref{fig::motivation_1}, regardless of the architecture of GNNs, both $\textrm{MLP}_\textrm{GNN}^{Share}$ and $\textrm{MLP}_\textrm{GNN}^{Re}$ exhibit inferior performance compared to the corresponding GNN. It demonstrates that the effective utilization of topology is one of the keys to the excellent performance of GNN, which is also consistent with the findings of previous studies \cite{SGC,deeper_insight}. 
Meanwhile, it also should be noticed that $\textrm{MLP}_\textrm{GNN}^{Share}$  outperforms $\textrm{MLP}_\textrm{GNN}^{Re}$ on the Cora and Citeseer datasets and $\capSh$ improves significantly compared to $\capRe$. 
It shows that the explicit structural bias brought by \textbf{P} operations may benefit $\textrm{MLP}_\textrm{GNN}$ to capture the structural information to some extent. 

Besides, as shown in Table~\ref{tab::Disassort}, 
% the \bfP utilizing the structure inherently introduces a strong relational induction bias in the representation learning, as pointed out in~\cite{battaglia2018relational}.
while \bfP facilitates the classification of assortative nodes for GNN, it also leads to difficulties in dealing with disassortative nodes. 
% since the introduced relational induction bias. 
On the contrary, when dealing with those disassortative nodes, $\textrm{MLP}_\textrm{GNN}^{Re}$ without using any structural prior shows a more powerful ability than GNN. 
% which generally copes well with assortative nodes containing strong structural relationships. 
% as a model that is perceived as having no or very little inductive bias~\cite{battaglia2018relational},
% MLP can perform representation learning by relying only on the raw features of nodes in the absence of structural prior.

More importantly, as we can observe from Fig.~\ref{fig::motivation_2}, $\cupRe$ shows the best performance on each dataset whatever the GNN is used. 
$\cupRe$ can outperform the GNN by about 5\%-10\%, which demonstrates that $\textrm{MLP}_\textrm{GNN}^{Re}$ captures some beneficial information that the GNN cannot grasp from the node content feature, despite the poor performance of $\textrm{MLP}_\textrm{GNN}^{Re}$. 
Besides, although $\cupSh$ declines slightly compared to $\cupRe$, it still outperforms the GNN by a considerable amount. This indicates that the \textbf{P} operation is also unable to help the host GNN retain the specific knowledge obtained by $\textrm{MLP}_\textrm{GNN}$ from node features.

\noindent\textbf{Summary.} 
The empirical investigation mentioned above can roughly provide an answer for (\textbf{Q1}) that GNN does not unleash the full potential of \bfT operations no matter what the architecture of the GNN is, since the GNN throws away the valuable information already captured by \bfT operations from node content feature. 
In addition, this case study also hints at an interesting observation for (\textbf{Q2}) that \bfP operations could inject some knowledge about the topological structure of the graph from the host GNN into $\textrm{MLP}_\textrm{GNN}$ to strengthen its capability. On the contrary, \bfP operations cannot realize the knowledge transfer from $\textrm{MLP}_\textrm{GNN}$ to the GNN as a whole. Furthermore, the \bfP operation inherently introduces a strong relational induction bias in the representation learning process, as pointed out in~\cite{battaglia2018relational}. It brings GNN the ability which generally cope well with assortative nodes containing strong structural relationships. 
As a model that is perceived as having no or very little inductive bias~\cite{battaglia2018relational},
MLP can perform representation learning by relying only on the raw features of nodes in the absence of structural prior. 
It means that the \textbf{T}-only training for $\textrm{MLP}_\textrm{GNN}$ can be favorably served as a complementary of \bfP operation of the host GNN, allowing us to achieve a better trade-off between node feature modeling and structural biases.

% As illustrated in Fig.~\ref{fig::motivation_framework}, these observations mentioned above motivate us to seek a mechanism to achieve the mutual knowledge transfer between the GNN and $\textrm{MLP}_\textrm{GNN}$, thereby improving the ability of feature transformation in GNNs and injecting the structure information into the feature transformation.

\section{Bi-directional Knowledge Transfer}
% After recognizing the potential of \bfT operation and the cross-influence between \bfT operation and \bfP operation, 
% Our goal is to establish an effective training mechanism for GNNs that seamlessly combines the advantages of feature transformation and feature propagation in aspects of feature modeling and structural inductive bias. 

As illustrated in Fig.~\ref{fig::motivation_framework}, The observations mentioned above motivate us to seek a mechanism to achieve the mutual knowledge transfer between the GNN and $\textrm{MLP}_\textrm{GNN}$, thereby seamlessly combining the advantages of feature transformation and feature propagation in aspects of feature modeling and structural inductive bias.

For this purpose, rather than introducing a new GNN by meticulously altering the model architecture, the proposed Bi-directional Knowledge Transfer (\bikit) accomplishes knowledge capture from both the GNN and $\MLPG$ by modeling their representation distributions, and then progressively infuses the knowledge into each other.

% In this section, we first present how to extract and transfer the model knowledge between the host GNN and $\textrm{MLP}_\textrm{GNN}$, and then we illustrate the alternate training process of GNN and $\textrm{MLP}_\textrm{GNN}$ with knowledge transfer. Finally, we theoretically demonstrate the feasibility of the method and its benefits for the generalization of GNN and the derived $\textrm{MLP}_\textrm{GNN}$.
% The overview of \bikit~is shown in Fig.~\ref{fig::overall_framework}.

% \begin{figure}[t]	
% 	\centering
% 	\includegraphics[width=0.48\textwidth]{Fig/BIKITV3.pdf}
% 	\caption{An illustration of \bikit~with a 3-layer GCN as an example. 
%  % We train GNN and $\textrm{MLP}_\textrm{GNN}$ alternately. 
%  % After GNN training is complete, 
%  $G_{gnn}$ learns to model distribution $q_{gnn}(\rvz|y)$ in accordance with the trained GNN.
%  Then, in the training of $\textrm{MLP}_\textrm{GNN}$, $G_{gnn}$ generates samples in the representation space to facilitate the training of $\textrm{MLP}_\textrm{GNN}$. 
%  Similarly, $G_{mlp}$ then learns to model $q_{mlp}(\rvz|y)$ according to the trained $\textrm{MLP}_\textrm{GNN}$ and is used in the next training of GNN.}
% 	\label{fig::overall_framework}
%  \vspace{-8pt}
% \end{figure}

\begin{figure*}
    \centering
    \includegraphics[width=\textwidth]{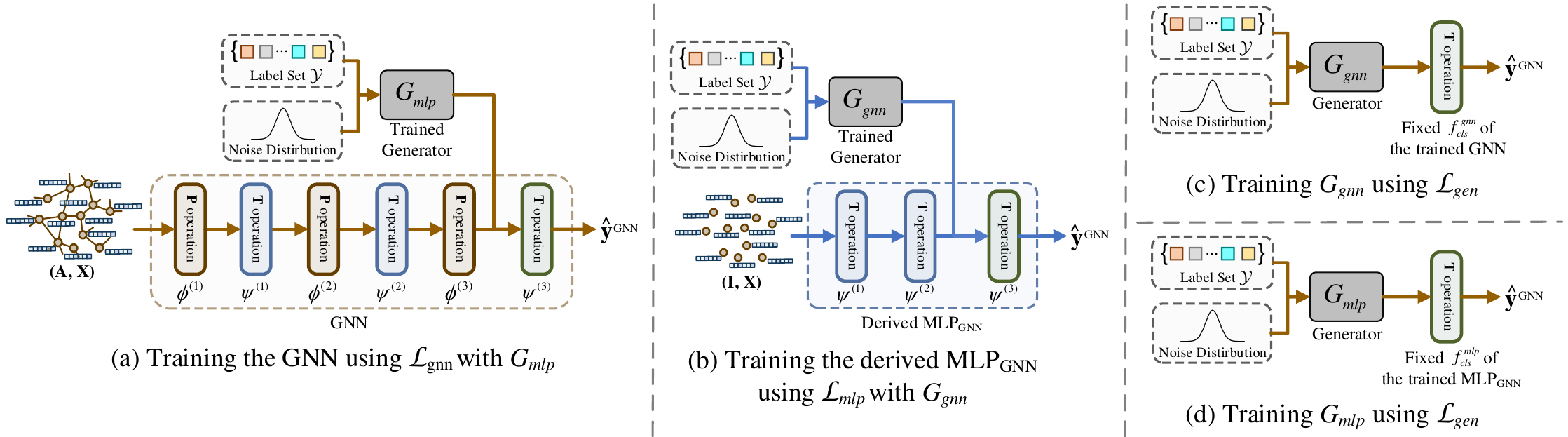}
    \caption{An overview of BiKT to show the parts of the training involved in each phase. During the recurrent training, (a)(c) and (b)(d) would be performed alternately to establish the bi-directional knowledge transfer between GNN and $\MLPG$.}
    \label{fig::training_detail}
\end{figure*}

\subsection{Generation-based Representation Distribution Modeling} 
% The representation distribution from GNN embeds the joint knowledge of structural information and node features, while the representation distribution from $\MLPG$ captures knowledge only from node features.  
Not limited to model parameters, a broader view of model knowledge is a learned mapping from input to output~\cite{hinton2015distilling}. 
Considering the structural bias introduced by \bfP in GNN during representation learning, to capture the model knowledge from GNNs and $\MLPG$, we utilize the representation distribution of the model as a surrogate for the learned model mapping to instantiate the model knowledge.

% To capture the model knowledge from GNNs and $\MLPG$, we instantiate the knowledge as the representation distribution of models. 

In particular, let $\gX  \subset \R^d$ be the raw feature space and $\gZ \subset \R^{d_m}$ be the representation space.  For a $L$-layer GNN, it can be separated into two components: a feature extractor $f_{gnn}(\rmX, \rmA): \gX \to \gZ$  composed by all operations except for $\psi^{(L)}$, and a classifier $f_{cls}(\rmZ): \gZ \to \gY$ played by $\psi^{(L)}$. Similarly, $\MLPG$ can also be separated as: $f_{mlp}(\rmX)$ composed by $\{\psi^{(l)}\}_{l=1}^{L-1}$, and $f_{cls}(\rmZ)$. 
Due to the explicit involvement of $\rmA$ in the representation learning process, the representation distribution modeled by \bfT operations has been altered by additional \bfP operations. It not only injects the information of topological structure into the representation distribution but also interferes with the knowledge from content features. It means the representation distribution from $f_{gnn}$ can be seen that embed the joint knowledge of structural information and node features, while the representation distribution from $f_{mlp}$ captures knowledge only from node features. 
% Concretely, the representation distribution extracted by $f_{mlp}$ only embeds the knowledge captured from the raw content feature of nodes. While for $f_{gnn}$, the representation distribution extracted by shared \bfT operations in it has altered due to the intervention of additional \bfP operations. It not only injects the information of topological structure into the representation distribution but also interferes with the knowledge from content features. 
% Therefore, the model knowledge of GNN and $\MLPG$ can be characterized by modeling their representation distribution respectively.

% Noting that although $f_{cls}$ is shared between the host GNN and $\MLPG$, it will adapt itself according to the input representation during the training of the host GNN and $\MLPG$ respectively, thus to learn the optimal mapping from $\gZ$ to $\gY$.  

% Concretely, the representation distribution extracted by $f_{mlp}$ only embeds the knowledge captured from the raw content feature of nodes. While for $f_{gnn}$, the representation distribution extracted by shared \bfT operations in it has altered due to the intervention of additional \bfP operations. It not only injects the information of topological structure into the representation distribution but also interferes with the knowledge from content features. 
% Therefore, it is difficult for the classifier $f_{cls}$ to coordinate the knowledge from the topological structure and content feature at the same time when participating in the training of GNN or $\MLPG$.

To achieve the representation distribution modeling effectively, we consider learning an auxiliary conditional distribution $q(\rvz|y):\gY \to \gZ$. It is applied to fit the representation distribution extracted by the given model, thus back-capturing the knowledge embedded in the representation distribution. 
% The benefits of this approach are manifested obviously in these aspects: 
% I)
Let $p(y|\rvz)$ be the learned posterior distribution of the model and $p(y)$ be the ground-truth prior distribution of labels, $q(\rvz|y)$ can be modeled as:
\begin{align}
\label{obj::knowledge-raw}
    {q(\rvz|y)} =  \argmax_{q(\rvz|y)} ~ H(q(\rvz|y))+\E_{y \sim p(y)} \E_{\rvz \sim q(\rvz|y)}[\log p(y|\rvz)]
\end{align}
where $H(q(\rvz|y))$ denotes the entropy of $q(\rvz|y)$, which is used to ensure the diversity of $\rvz$.

Specifically, we adopt the uniform distribution $\hat{p}(y)$ as an alternative to $p(y)$, since not all nodes' labels are available and therefore $p(y)$ is unknown during the training phase. The counting of the observable labels in the training set is also a common practice to approximate $p(y)$.
Further, inspired by the idea of generative models~\cite{VAE,GAN,LDM}, a conditional generator $G$ is adopted to learn $q(\rvz|y)$ as $G(y, \varepsilon   | \varepsilon   \sim \gN(0, \rmI_{d_m}))$, where $\varepsilon  $ is a noise vector sampled from $d_m$-dimensional standard normal distribution. Since we rely on the generator $G$ to realize the non-parametric estimation of $q(\rvz|y)$, it is not easy to calculate $H(q(\rvz|y))$ directly. To bypass this problem, the mode-seeking regularization term $D(G)$ from~\cite{mode_seeking} is adopted as a substitute for $H(q(\rvz|y))$ to escort the diversity of $\rvz$:
\begin{align}
\label{obj::mode_seeking}
    % D(G) = \E_{y \sim p(y)} \E_{ \varepsilon  _1 \sim \gN(0, \rmI_{d_m})} \E_{\varepsilon  _2 \sim \gN(0, \rmI_{d_m})}[\frac{d_{\rvz}(G(y,\varepsilon  _1), G(y,\varepsilon  _2))}{d_{\varepsilon  }(\varepsilon  _1, \varepsilon  _2)}]
    {D(G)} = \max_{G} \E_{y \sim p(y)} \E_{ \varepsilon  _1,\varepsilon  _2 \sim \gN(0, \rmI_{d_m})} [\frac{d_{\rvz}\left (G(y,\varepsilon  _1), G(y,\varepsilon  _2)\right )}{d_{\varepsilon  }(\varepsilon  _1, \varepsilon  _2)}]
\end{align}
where $d_{*}(\cdot)$ denotes the distance metric. According to Eq.~\eqref{obj::knowledge-raw} and Eq.~\eqref{obj::mode_seeking}, the generator $G$ can be trained to model $q(\rvz|y)$ through minimizing the following optimization objective:
\begin{equation}
\begin{aligned}
\label{obj::loss_G}
    % D(G) = \E_{y \sim p(y)} \E_{ \varepsilon  _1 \sim \gN(0, \rmI_{d_m})} \E_{\varepsilon  _2 \sim \gN(0, \rmI_{d_m})}[\frac{d_{\rvz}(G(y,\varepsilon  _1), G(y,\varepsilon  _2))}{d_{\varepsilon  }(\varepsilon  _1, \varepsilon  _2)}]
    \gL_{gen} = & \E_{y \sim \hat{p}(y)} \E_{ \varepsilon  \sim \gN(0, \rmI_{d_m})} \left [ \gL_{sl}(\sigma(f_{cls}(G(y,\varepsilon  ))), y) \right ]\\
    &- D(G)
\end{aligned}
\end{equation}
where $\gL_{sl}$ and $\sigma$ are the classification loss function and the activation function, respectively. It can be seen that we only need the classifier $f_{cls}$ to participate in the modeling of $q(\rvz|y)$, which is an efficient practice to capture the knowledge of models. 

% we could induce the corresponding representations learned by the model with given labels. These representations with model knowledge can be used in the training of other models to convey the knowledge between models. 
According to the generator learning mentioned above, \bikit~could acquire model knowledge from both the host GNN and the derived $\MLPG$ by fitting the representation distributions captured from them according to the two generators as shown in Fig.~\ref{fig::training_detail}(c) and (d), respectively.

\subsection{Knowledge Infusion} 
To achieve knowledge transfer between the GNN and $\MLPG$, we need to address how to inject knowledge gained from one model (target model) into the other model (source model).

To this end, we incorporate the generator $G_{tgt}$ into the training process of the source model as a regularizer, which has been trained to model the distribution $q_{tgt}(\rvz|y)$ of the target model according to Eq.~\eqref{obj::loss_G}. 
It means that the source model will be subject to two key constraints during the optimization process: the direct supervised loss from the downstream task, and the knowledge infusion loss from the generator.
% Taking the node classification task as an example, we present the details of knowledge infusion.
To be specific, given a label set $\{ \tilde{y}_{i}\}_{i=1}^K$ that sampled from $p(y)$, the corresponding representation set $\tilde\rmZ_{tgt} = \{ \tilde{\rvz}_{i}\}_{i=1}^K$ can be sampled from $q_{tgt}(\rvz|y)$ by $G_{tgt}$. Consequently, for the source model, the knowledge infusion regularization term can be formulated as:
 \begin{align}
\label{obj::loss_ki}
    \gL_{ki}(\tilde\rmZ_{tgt}, f_{cls}^{tgt}) = \sum_{\tilde{\rvz}_i \in \tilde\rmZ_{tgt} }\gL_{sl}(\sigma(f_{cls}^{tgt}(\tilde{\rvz}_i)), \tilde{y}_{i})
\end{align}
where $\gL_{sl}$ denotes the supervised loss, for the node classification task, we employ the cross-entropy loss as $\gL_{sl}$.

% Taking the node classification task as an example,
% 当我们把GNN作为source model， MLP作为target model时， balabala。 给MLP的部分也加上类似的话。
According to Eq.~\eqref{obj::loss_ki}, the generators can be involved in the training procedure of each model. 
Concretely,
when GNN serves as the source model and $\MLPG$ as the target model, GNN can integrate the knowledge from $\MLPG$ with the assistance of $G_{mlp}$ by minimizing the following optimization objective:
 \begin{align}
\label{obj::loss_GNN}
    % D(G) = \E_{y \sim p(y)} \E_{ \varepsilon  _1 \sim \gN(0, \rmI_{d_m})} \E_{\varepsilon  _2 \sim \gN(0, \rmI_{d_m})}[\frac{d_{\rvz}(G(y,\varepsilon  _1), G(y,\varepsilon  _2))}{d_{\varepsilon  }(\varepsilon  _1, \varepsilon  _2)}]
    \gL_{gnn} = \sum_{u \in \gV^{trn}}\gL_{sl}(\hat{\rvy}_u^{gnn}, y_u)  
    + \alpha \gL_{ki}(\tilde\rmZ_{mlp}, f_{cls}^{mlp})
\end{align}
where $\gV^{trn}$ denotes the set of training nodes,
$\hat{\rvy}^{gnn}_u$ denotes the predicted probability vector for node $u$ by the GNN, and $\alpha$ is a coefficient to control the strength of $\gL_{ki}$.

On the contrary, when $\MLPG$ serves as the source model and GNN as the target model, the total loss function for $\MLPG$ through \bikit~can be formulated as:
\begin{equation}
 \begin{aligned}
\label{obj::loss_MLP}
    % D(G) = \E_{y \sim p(y)} \E_{ \varepsilon  _1 \sim \gN(0, \rmI_{d_m})} \E_{\varepsilon  _2 \sim \gN(0, \rmI_{d_m})}[\frac{d_{\rvz}(G(y,\varepsilon  _1), G(y,\varepsilon  _2))}{d_{\varepsilon  }(\varepsilon  _1, \varepsilon  _2)}]
     \gL_{mlp} = \sum_{u \in \gV^{trn}}\gL_{sl}(\hat{\rvy}_u^{mlp}, y_u)  
    + &\alpha \gL_{ki}(\tilde\rmZ_{gnn}, f_{cls}^{gnn}) \\
    % + &\beta \sum_{u \in \gV} \gL_{ps}(\hat{\rvy}_u^{mlp}, \hat{\rvy}^{gnn}_u)
    + &\beta \gL_{ps}(\hat{\rmY}_{\gV}^{mlp}, \hat{\rmY}^{gnn}_{\gV})
\end{aligned}
\end{equation}
where $\hat{\rvy}^{mlp}_u$ denotes the predicted probability vector for node $u$ by $\MLPG$ and the KL-divergence is used to serve as $\gL_{ps}$. Besides, $\hat{\rmY}_{\gV}^{*}=\{\hat{\rvy}^{*}_u\}_{u\in\gV}$ denotes the set of the predicted probability vector for $\gV$, $\hat{\rmY}_{\gV}^{mlp}$ and $\hat{\rmY}_{\gV}^{gnn}$ is the sets corresponding to $\MLPG$ and GNN respectively, $\beta$ is also a coefficient to control the strength of $\gL_{ps}$. 

Compared to Eq.~\eqref{obj::loss_GNN} for GNN, it can be observed from Eq.~\eqref{obj::loss_MLP} includes an additional term $\gL_{ps}$. This regularization term is introduced to reduce the huge gap in performance between $\MLPG$ and GNN at the initial stage. $\gL_{ps}$ utilizes the predictions of GNN as reliable results to guide the training of $\MLPG$. This provides a direct and efficient method for transferring structural information from GNN to $\MLPG$, complementing the indirect distribution-level knowledge infusion through $G_{gnn}$.

% 这个地方我们先把公式7给出来，已和公式6形成对照。完事再去解释公式7增加了什么。
% Considering the large performance gap between $\MLPG$ and GNN in the initial phase, for \bikit-$\MLPG$, we also employ the prediction of GNN as a trustable result to guide the training of $\MLPG$. 
% % in order to directly and quickly migrate structural information to $\MLPG$. 
% Compared to the indirect distributional-level knowledge infusion through $G_{gnn}$, it serves as a direct way that efficiently migrate structural information from GNN to $\MLPG$. 

\begin{figure}
    \centering
    \includegraphics[width=0.5\textwidth]{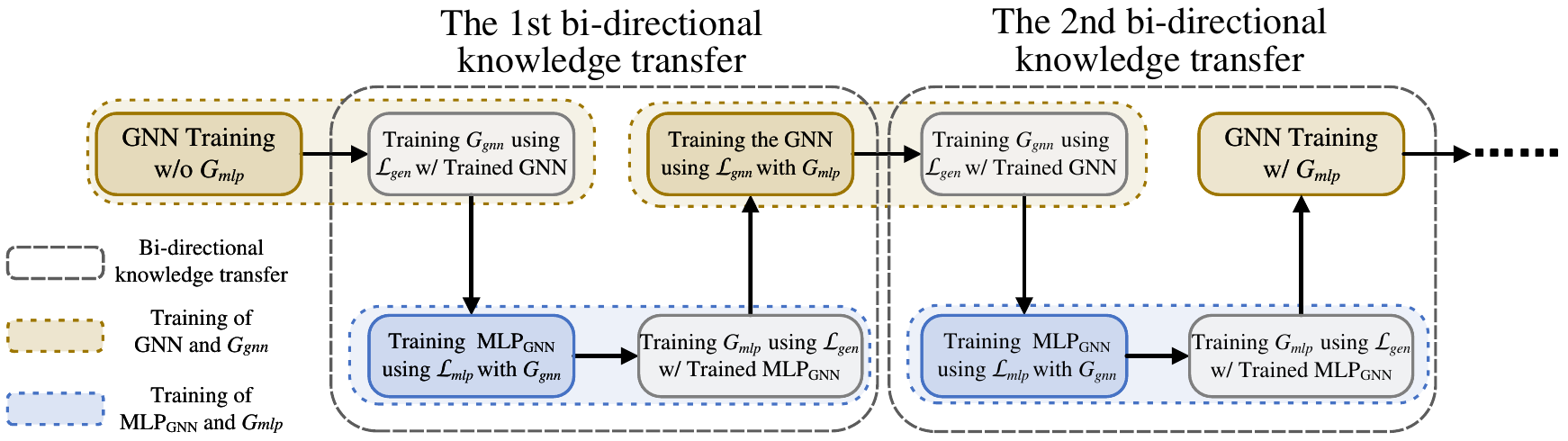}
    \caption{The illustration of the recurrent training with parameter inheritance between $\MLPG$ and GNN. This practice facilitates the progressive knowledge integration of both sides.}
    \label{fig::example}
\end{figure}

\subsection{Progressive Knowledge Transfer with Parameter Inheritance} 
% 先描述一下为什么非要做渐进式知识迁移。再具体展开。

% A straightforward intuition is that transferring all knowledge during a single training process can be challenging for the model. Meanwhile, the host GNN can only benefit from the initial $\MLPG$, not from the optimized $\MLPG$ according to Eq.~\eqref{obj::loss_MLP}. 
% % To bridge the disparity in knowledge between GNN and $\MLPG$ and harness the complementary of the \textbf{T} operations within both, 
% Therefore, we adopt a progressive transfer strategy between $\MLPG$ and GNN to integrate knowledge from both sides.
Generally speaking, a one-way knowledge transfer from $\MLPG$ to the host GNN can achieve the goal of enhancing \bfT within GNN. However, such a practice only allows the host GNN to benefit from the initial $\MLPG$, not from the optimized $\MLPG$ according to Eq.~\eqref{obj::loss_MLP}. Moreover, \bfT still relies on \bfP to explicitly inject structural bias and cannot autonomously learn this inductive bias by itself. 
Inspired by the concept of continuous learning~\cite{wang2023comprehensive}, we adopt a recurrent training strategy between $\MLPG$ and GNN to integrate knowledge from both sides progressively.

To be specific, the host GNN and $\MLPG$ are trained alternately in the form of recurrent training with parameter inheritance, as shown in Fig.~\ref{fig::example}.
The model parameter is a foundational form of model knowledge~\cite{hinton2015distilling}, which can be shared directly between models of the same architecture to enable knowledge inheritance. 
Moreover, it can be observed from Section~\ref{sect::empirical} that the parameters of a properly trained GNN provide a favorable initialization for $\MLPG$ compared to a random initialization. 
Hence, for the host GNN and the derived $\MLPG$, after either model completes the training procedure, its model parameters are directly inherited by the other model as model initialization. This strategy not only facilitates knowledge transfer between models, but also alleviates knowledge forgetting that may occur during training.

% Moreover, parameter sharing between GNN and $\MLPG$ can be seen as parameter inheritance between the two, facilitating knowledge transfer.
% Drawing inspiration from these insights, we adopt an iterative training strategy with parameter inheritance between $\MLPG$ and GNN to progressively integrate knowledge from both sides.
% Benefiting from parameter inheritance and knowledge infusion, the knowledge between GNN and $\MLPG$ during recurrent training can be progressively integrated rather than forgotten.
Based on the parameter inheritance, the host GNN and $\MLPG$ can be trained alternately based on each other's parameter knowledge. 
We alternate the roles of the host GNN and $\MLPG$ as the source model and the target model during the recurrent training process to establish bi-directional knowledge transfer. This process shares a fascinating similarity indeed with the classical co-training paradigm~\cite{co-training}.
In practice, we first perform a standard training of the host GNN to obtain a base model according to Eq.~\eqref{obj::loss_GNN} with $\alpha=0$. Then, a complete bi-directional knowledge transfer between the host GNN and the derived $\MLPG$ can be described as follows:
 Based on the trained $f_{cls}^{gnn}$ in the host GNN, a generator $G_{gnn}$ is instantiated to model $q_{gnn}(\rvz|y)$ according to Eq.~\eqref{obj::loss_G}. Next, according to Eq.~\eqref{obj::loss_MLP} and $G_{gnn}$, we proceed with the further training of the $\MLPG$ which inherits parameters from the host GNN. After the training of $\MLPG$, $G_{mlp}$ is used to learn $q_{mlp}(\rvz|y)$. Similarly, in the next training phase of the host GNN, $G_{mlp}$ can play a vital role in seamlessly integrating the knowledge, which exclusively stems from node features via $\MLPG$, into the host GNN according to Eq.~\eqref{obj::loss_GNN}. 
% This constitutes a complete knowledge transfer, as shown in Fig.~\ref{fig::example}. 
 The process can be executed recurrently several times according to the parameter inheritance mechanism.

\subsection{Generalization Analysis of \bikit}
The core idea of \bikit~is to extract model knowledge from the host GNN and the derived $\MLPG$ by modeling the representation distribution $q_{gnn}(\rvz|y)$ and $q_{mlp}(\rvz|y)$ and then integrating the distribution into the learning process of each other to achieve knowledge transfer. As a result, \textbf{\textit{whether the generator can successfully fit the induced distribution of the target model from $\gY$ to $\gZ$}} plays a crucial role in the feasibility of \bikit. To dispel such doubt, we have the following proposition:
\begin{proposition}
\label{thm::distribution_matching}
Let $q(\rvz|y)$ be the distribution modeled by generator $G$, $p(y|\rvz;f_{cls})$ and $p(\rvz|y;f_{cls})$ be the posterior distribution and the corresponding induce distribution of $f_{cls}$. Then maximizing Eq.~\eqref{obj::knowledge-raw} is equivalent to minimizing the conditional KL-divergence between $q(\rvz|y)$ and $p(\rvz|y;f_{cls})$.
\end{proposition}
\noindent For proof of \textbf{Proposition}~\ref{thm::distribution_matching}, please refer to the Appendix\ref{appendix::proof}. 

By showing the connection between the optimization objective of the generator and the conditional KL-divergence between $q(\rvz|y)$ and $p(\rvz|y;f_{cls})$, \textbf{Proposition}~\ref{thm::distribution_matching} demonstrates that generator could fit the induced distribution of the target model based on the optimization objective. 

Moreover, as we mentioned in Sect.~\ref{Sect::Pre}, the host GNN is equivalent to $\MLPG$ in the case where $\rmA$ is set to an identity matrix $\rmI_{n}$. Therefore, by treating the node features $\rmX$ and topological structure $\rmA$ as a whole set,
we could define the distributions of set $(\rmX, \rmA)$ and $(\rmX, \rmI_{n})$ as $\gD_{gra}$ and $\gD_{fea}$ respectively. 
% Then, on the basis of the definition from~\cite{DA} to see the distributions $\gD_{gra}$ and $\gD_{fea}$ as two domains, the problem studied in this paper can be regarded as a domain adaption problem. Specifically, after we finish training the host GNN, we test the performance of the $\MLPG$ in it, which is equivalent to testing the effect of migrating the model to domain $\gD_{fea}$ after we finish training in domain $\gD_{gra}$.
Following the definition provided in~\cite{DA}, we can view the distributions $\gD_{gra}$ and $\gD_{fea}$ as two distinct domains. The problem addressed in this paper can thus be framed as a domain adaptation problem. To elaborate, once we have completed the training of the host GNN, evaluating the performance of the $\textrm{MLP}_\textrm{GNN}^{Share}$ is equivalent to assessing the impact of transferring the model to the $\gD_{fea}$ domain after training in $\gD_{gra}$.
Therefore, we can analyze \textbf{\textit{whether the source model can benefit from the captured distribution of the generator}} from the perspective of domain adaption.
We first let the distribution $q(\rvz|y)$ derived by the generator satisfy the following assumption after optimization according to \textbf{Proposition}~\ref{thm::distribution_matching}:
\begin{assumption}
\label{thm::assumption}
 % Given $f_{cls}$ of the target model, $q(\rvz|y)$ and $p(\rvz|y;f_{cls})$ can be optimally matched, i.e., $D_{KL}[q(\rvz|y)||p(\rvz|y;f_{cls})]=0$.
 Given $f_{cls}^{tgt}$ and $f_{cls}^{src}$ of the target model and source model respectively, $q(\rvz|y)$ approximate $p(\rvz|y;f_{cls}^{tag})$ and have $d_{\gH} (q(\rvz|y), p(\rvz|y;f_{cls}^{tag}))  \leq d_{\gH} (p(\rvz|y;f_{cls}^{src}), p(\rvz|y;f_{cls}^{tag}))$, where $d_{\gH}$ denotes the $\gH$-divergence.
\end{assumption}
Then, based on the generalization bound for domain adaption proposed in~\cite{DA, DA2}, we have: 

\begin{proposition}
\label{thm::generalization}
Suppose assumption 1 holds. Let $\gT$ and $\gS$ be the source domain and target domain with the distribution $\gD_{s}$ and $\gD_{t}$, respectively. Let $\gR$ be a representation function from $\gX$ to $\gZ$. 
Denote $\gD_{G}$ be an auxiliary distribution derived from a generator $G$ 
and $\gD_{s}^{'} = \tau \gD_{G} + (1-\tau) \gD_{s}$.
Denote $\gH$ be a set of hypothesis with VC-dimension $d$. Given an empirical dataset $\hat{\gD}_{s}$ and an augmented dataset $\hat{\gD}_{s}^{'}$ with $|\hat{\gD}_{s}|=m$ and $|\hat{\gD}_{s}^{'}|=m' > m$. 
Let $J(m,d) = \sqrt{\frac{4}{m} \left( d \log\frac{2 e m }{d} + \log \frac{4}{\delta} \right) }$, where $e$ is the base of the natural logarithm.
If $(\epsilon_{\gS}(h) - \epsilon_{\gS^{'}}(h))$ is bounded and $m > \frac{1}{2}d$, then with probability at least $1-\delta$, for every hypothesis $h \in \gH$:
\begin{equation}
    \begin{aligned} 
    % \epsilon_{\gT}(h) 
    % &\leq \varepsilon_{\gS'}(h) 
    % + \sqrt{\frac{4}{m'} \left( d \log\frac{2 e m' }{d} + \log \frac{4}{\delta} \right) }  
    %  %
    %  + d_{\gH} (\Dz_{s}^{'}, \Dz_{t}) + \lambda' \\
    %  & \leq \epsilon_{\gS}(h)  + \sqrt{\frac{4}{m} \left( d \log\frac{2 e m }{d} + \log \frac{4}{\delta} \right) } + d_{\gH} (\Dz_{S}, \Dz_{T}) + \lambda
    %%%%%%%%%%%%%%%%%%%%%%%%%%%%%%%%%%%%%%%%%%%%%%%%%%%%%%%%%%%%%%%%%%%%%
    \epsilon_{\gT}(h) 
    & \leq \hat{\epsilon}_{\gS'}(h) 
    + J(m',d)
     + d_{\gH} (\Dz_{s}^{'}, \Dz_{t}) + \lambda' \\
     & \leq \hat{\epsilon}_{\gS}(h)  
     + J(m,d)
     + d_{\gH} (\Dz_{s}, \Dz_{t}) + \lambda
    \end{aligned}
\end{equation}
where $\epsilon_{*}(h)$ and $\hat{\epsilon}_{*}(h)$ are the expected and empirical risk of $h$ on the domain respectively. $\gS'$ denotes the augmented source domain with $\Dz_{s}^{'}$, $\lambda = \underset{h}{\min} \left( \epsilon_{\gT}(h) + \epsilon_{\gS}(h) \right )$ and $\lambda' = \underset{h}{\min} \left( \epsilon_{\gT}(h) + \epsilon_{\gS'}(h) \right )$ is the optimal risk on two domains.
\end{proposition}

\begin{table*}[t]
\centering
\caption{Results of the node classification task under the transductive setting, where \tcr{red} letters denote the performance gains brought by \bikit~compared to the base $\textrm{MLP}_\textrm{GNN}$ and $\textrm{GNN}$, respectively.}
\scriptsize
\begin{tabular}{@{}c|ccccc|cc@{}}
\toprule
Method         & Cora & Citeseer & Pubmed & A-computer & A-photo & Chameleon & Squirrel \\ \midrule

$\textrm{MLP}_\textrm{GCN}$ 
& 49.84$\pm$2.19
& 53.77$\pm$2.13
& 79.22$\pm$0.77
& 72.69$\pm$0.91
& 79.51$\pm$1.67
& 27.60$\pm$2.17
& 22.83$\pm$1.14
\\

\bikit-$\textrm{MLP}_\textrm{GCN}$ 
&  74.99$\pm$2.54  
&  69.61$\pm$1.55  
&  85.15$\pm$0.93  
&  81.26$\pm$1.74  
&  90.58$\pm$1.05       
&  39.69$\pm$3.03
&  31.42$\pm$2.39
\\

GCN     
&  73.44$\pm$1.96  
&  65.97$\pm$2.46  
&  81.62$\pm$0.90  
&  83.47$\pm$1.34  
&  90.10$\pm$0.74       
&  39.22$\pm$3.59
&  30.44$\pm$2.09
\\

\bikit-GCN 
&  76.91$\pm$2.87  
&  68.63$\pm$2.06  
&  82.28$\pm$0.46  
&  84.24$\pm$1.05  
&  90.90$\pm$0.74       
&  41.48$\pm$2.53
&  31.19$\pm$1.87
\\  

\textbf{Improv.} (MLP/GNN) 
&\tcr{25.15}/\tcr{3.47}
&\tcr{15.84}/\tcr{2.66}
&\tcr{5.93}/\tcr{0.66}
&\tcr{8.57}/\tcr{0.77}
&\tcr{11.07}/\tcr{0.80}
&\tcr{12.09}/\tcr{2.26}
&\tcr{8.59}/\tcr{0.75} \\ \midrule

% &  \tcr{1.55}/\tcr{3.47}   
% &  \tcr{3.64}/\tcr{2.66}   
% &  \tcr{3.53}/\tcr{0.66}   
% &  \tccy{-2.21}/\tcr{0.77}   
% &  \tcr{0.48}/\tcr{0.80}   \\ \midrule

$\textrm{MLP}_\textrm{GAT}$ 
& 47.80$\pm$2.41
& 49.41$\pm$1.95
& 79.61$\pm$0.72
& 69.68$\pm$1.09
& 74.60$\pm$1.60
& 39.83$\pm$2.07
& 22.78$\pm$1.58
\\

\bikit-$\textrm{MLP}_\textrm{GAT}$ 
&  71.60$\pm$3.16  
&  60.79$\pm$2.33  
&  82.53$\pm$0.72  
&  82.68$\pm$1.47  
&  87.51$\pm$2.98       
&  41.34$\pm$2.89
&  29.77$\pm$1.75
\\

GAT     
&  72.67$\pm$3.80  
&  64.43$\pm$3.08  
&  81.32$\pm$0.71  
&  83.42$\pm$1.68  
&  88.14$\pm$2.08       
&  41.27$\pm$2.58
&  27.01$\pm$1.71
\\

\bikit-GAT 
&  77.40$\pm$1.50  
&  67.11$\pm$1.70  
&  81.91$\pm$0.54  
&  84.48$\pm$1.17  
&  90.65$\pm$0.54
&  42.89$\pm$3.91
&  27.69$\pm$1.77
\\

\textbf{Improv.} (MLP/GNN)  
& \tcr{23.80}/\tcr{4.73}  
& \tcr{11.38}/\tcr{2.68}  
& \tcr{2.92}/\tcr{0.59}
& \tcr{13.00}/\tcr{1.06}  
& \tcr{12.91}/\tcr{2.51}      
& \tcr{1.51}/\tcr{1.62}
& \tcr{6.99}/\tcr{0.68}
\\ \midrule

$\textrm{MLP}_\textrm{FAGCN}$ 
& 48.77$\pm$2.64
& 52.16$\pm$2.82
& 80.13$\pm$0.76
& 70.14$\pm$1.19
& 78.29$\pm$2.28
& 31.78$\pm$6.92
& 22.96$\pm$2.43
\\

\bikit-$\textrm{MLP}_\textrm{FAGCN}$ 
&  73.64$\pm$2.98  
&  68.21$\pm$1.72  
&  82.83$\pm$1.14  
&  78.68$\pm$1.95  
&  89.31$\pm$1.30       
&  39.70$\pm$4.42
&  26.62$\pm$2.05
\\

FAGCN     
&  77.23$\pm$1.70  
&  67.47$\pm$1.71  
&  83.33$\pm$0.92  
&  83.19$\pm$1.25  
&  91.03$\pm$1.20       
&  41.82$\pm$3.94
&  27.09$\pm$1.96
\\

\bikit-FAGCN 
&  79.00$\pm$1.65  
&  68.79$\pm$3.10  
&  84.14$\pm$0.61  
&  84.49$\pm$0.97  
&  92.16$\pm$0.28       
&  43.07$\pm$3.43
&  28.10$\pm$1.65
\\

\textbf{Improv.} (MLP/GNN)  
&\tcr{24.87}$\pm$\tcr{1.77}
&\tcr{16.05}$\pm$\tcr{1.32}
&\tcr{2.70 }$\pm$\tcr{0.81}
&\tcr{8.54}$\pm$\tcr{1.30}
&\tcr{11.02}$\pm$\tcr{1.13}
&\tcr{7.92}$\pm$\tcr{1.25}
&\tcr{3.66}$\pm$\tcr{1.01}

\\ \midrule

$\textrm{MLP}_\textrm{GCNII}$ 
& 49.75$\pm$2.69
& 52.22$\pm$2.53
& 79.95$\pm$0.75
& 71.28$\pm$1.10
& 78.97$\pm$1.59
& 37.19$\pm$2.45
& 25.00$\pm$1.80
\\

\bikit-$\textrm{MLP}_\textrm{GCNII}$ 
&  75.72$\pm$2.72  
&  68.92$\pm$2.17  
&  86.05$\pm$0.52  
&  84.39$\pm$0.91  
&  92.25$\pm$1.00       
&  41.69$\pm$2.90
&  30.05$\pm$1.33
\\

GCNII     
&  79.45$\pm$0.92  
&  68.36$\pm$1.63  
&  83.98$\pm$1.09  
&  84.82$\pm$1.00  
&  90.64$\pm$1.73         
&  40.26$\pm$2.85
&  27.85$\pm$2.70
\\

\bikit-GCNII 
&  80.19$\pm$1.40  
&  69.35$\pm$1.81  
&  85.09$\pm$0.40  
&  85.40$\pm$1.07  
&  91.93$\pm$0.75       
&  42.78$\pm$3.30
&  29.11$\pm$1.67
\\

\textbf{Improv.} (MLP/GNN)  
&\tcr{25.97}$\pm$\tcr{0.74}
&\tcr{16.70}$\pm$\tcr{0.99}
&\tcr{6.10}$\pm$\tcr{1.11}
&\tcr{13.11}$\pm$\tcr{0.58}
&\tcr{13.28}$\pm$\tcr{1.29}
&\tcr{4.50}$\pm$\tcr{2.52}
&\tcr{5.05}$\pm$\tcr{1.26}

\\ \midrule

$\textrm{MLP}_\textrm{MixHop}$ 
& 48.04$\pm$3.93
& 49.28$\pm$2.87
& 79.48$\pm$1.15
& 71.30$\pm$1.31
& 78.17$\pm$2.15
& 27.46$\pm$4.28
& 21.78$\pm$0.84
\\

\bikit-$\textrm{MLP}_\textrm{MixHop}$ 
&  77.49$\pm$1.15  
&  68.11$\pm$1.78  
&  85.55$\pm$0.72  
&  82.96$\pm$1.15                 
&  91.26$\pm$1.16       
&  34.93$\pm$3.85
&  29.70$\pm$2.32
\\

MixHop     
&  74.60$\pm$1.96  
&  62.64$\pm$2.26  
&  84.62$\pm$0.65  
&  78.98$\pm$1.23  
&  87.44$\pm$1.58       
&  37.52$\pm$7.07
&  27.55$\pm$3.53
\\

\bikit-MixHop 
&  79.43$\pm$2.29  
&  68.28$\pm$1.55  
&  85.68$\pm$0.73  
&  83.58$\pm$1.17                 
&  91.44$\pm$1.40       
&  39.24$\pm$5.27
&  28.91$\pm$1.78
\\

\textbf{Improv.} (MLP/GNN)  
&\tcr{29.45}$\pm$\tcr{4.83}
&\tcr{18.83}$\pm$\tcr{5.64}
&\tcr{6.07}$\pm$\tcr{1.06}
&\tcr{11.66}$\pm$\tcr{4.60}    
&\tcr{13.09}$\pm$\tcr{4.00}
&\tcr{7.47}$\pm$\tcr{1.72}
&\tcr{7.92}$\pm$\tcr{1.36}

\\ \bottomrule
\end{tabular}
\label{tab:transductive_main_reuslts}
\vspace{-10pt}
\end{table*}

\begin{table*}[t]
\centering
\scriptsize
\caption{Results of the node classification task under the inductive setting, where \tcr{red} letters denote the performance gains brought by \bikit~compared to the base $\textrm{MLP}_\textrm{GNN}$ and $\textrm{GNN}$, respectively.}
\begin{tabular}{@{}c|ccccc|cc@{}}
\toprule
Method         
& Cora 
& Citeseer 
& Pubmed 
& A-computer 
& A-photo 
& Chameleon 
& Squirrel \\ \midrule

%MLP                 
%&  51.90\tiny{$\pm$2.25}  
%&  53.04\tiny{$\pm$6.54}  
%&  79.37\tiny{$\pm$0.87}  
%&  78.77\tiny{$\pm$1.29}  
% &  71.87\tiny{$\pm$1.54}      \\ \midrule

$\textrm{MLP}_\textrm{GCN}$       
&  54.47$\pm$3.31 
&  58.54$\pm$2.43 
&  76.47$\pm$3.44 
&  58.21$\pm$6.51 
&  79.42$\pm$2.69
&  26.31$\pm$2.00
&  24.27$\pm$1.63
\\

\bikit-$\textrm{MLP}_\textrm{GCN}$      
&  65.70$\pm$3.14  
&  65.52$\pm$3.97  
&  84.19$\pm$1.13  
&  79.54$\pm$0.94  
&  87.15$\pm$1.01
& 38.72$\pm$3.21
& 30.14$\pm$2.74
\\ 

GCN       
&  75.18$\pm$1.73  
&  65.32$\pm$2.01  
&  83.71$\pm$1.30  
&  84.26$\pm$1.55  
&  91.11$\pm$0.69
& 38.97$\pm$3.12
& 30.09$\pm$2.62
\\

\bikit-$\textrm{GCN}$      
&  76.60$\pm$1.72  
&  67.20$\pm$2.41  
&  84.19$\pm$0.45  
&  85.92$\pm$1.26  
&  92.03$\pm$0.76
&  40.12$\pm$2.25
&  30.55$\pm$2.52
\\ 

\textbf{Improv.} (MLP/GNN)
& \tcr{11.23}/\tcr{1.42} 
& \tcr{6.98}/\tcr{1.88}
& \tcr{7.72}/\tcr{0.48}
& \tcr{21.33}/\tcr{1.66}
& \tcr{7.73}/\tcr{0.92}
& \tcr{12.41}/\tcr{1.15}
& \tcr{5.87}/\tcr{0.46}
\\ \midrule

% \textbf{Improv.}  
% &  \tcr{13.80}     
% &  \tcr{12.48}     
% &  \tcr{4.82}      
% &  \tcr{0.73}      
% &  \tcr{15.28}   \\ \midrule

$\textrm{MLP}_\textrm{GAT}$      
&  49.75$\pm$5.93 
&  51.33$\pm$4.54 
&  79.53$\pm$1.09 
&  57.39$\pm$12.4 
&  64.82$\pm$12.5
& 39.34$\pm$2.46
& 23.09$\pm$1.16
\\

\bikit-$\textrm{MLP}_\textrm{GAT}$      
&  62.20$\pm$3.47  
&  59.29$\pm$2.07  
&  82.33$\pm$0.63  
&  79.37$\pm$2.47  
&  84.76$\pm$1.66
& 40.55$\pm$3.20
& 29.28$\pm$1.36
\\

GAT     
&  72.55$\pm$2.80  
&  63.14$\pm$2.32  
&  81.46$\pm$0.81  
&  80.50$\pm$1.31  
&  84.14$\pm$1.60
& 40.06$\pm$3.07
& 27.73$\pm$1.13
\\

\bikit-GAT 
&  75.40$\pm$0.66  
&  66.39$\pm$1.21  
&  81.81$\pm$0.66  
&  83.23$\pm$2.03  
&  89.62$\pm$1.54
& 42.16$\pm$3.46
& 27.89$\pm$1.34
\\ 

\textbf{Improv.} (MLP/GNN)
&\tcr{12.45}/\tcr{2.85}
&\tcr{7.96}/\tcr{3.25}
&\tcr{2.80}/\tcr{0.35}
&\tcr{21.98}/\tcr{2.73}
&\tcr{19.94}/\tcr{5.48}
& \tcr{1.21}/\tcr{2.10}
& \tcr{6.19}/\tcr{0.16}
\\ \midrule

% \textbf{Improv.}  
% &  \tcr{10.30}     
% &  \tcr{6.25}      
% &  \tcr{2.86}      
% &  \tcr{0.60}      
% &  \tcr{12.89}       \\ 

$\textrm{MLP}_\textrm{FAGCN}$      
&  60.72$\pm$1.84  
&  58.84$\pm$2.46  
&  80.47$\pm$0.61  
&  69.16$\pm$2.44  
&  77.42$\pm$2.56       
& 31.25$\pm$6.39
& 23.45$\pm$2.26
\\

\bikit-$\textrm{MLP}_\textrm{FAGCN}$      
&  67.10$\pm$3.00  
&  65.49$\pm$2.13  
&  82.72$\pm$0.49  
&  77.57$\pm$1.31  
&  88.12$\pm$1.50       
& 40.28$\pm$4.60
& 26.26$\pm$2.49
\\

FAGCN     
&  73.19$\pm$3.58  
&  68.21$\pm$3.00  
&  82.49$\pm$1.92  
&  84.20$\pm$0.85  
&  92.13$\pm$2.10
& 41.57$\pm$4.25
& 27.70$\pm$1.81
\\

\bikit-FAGCN 
&  76.00$\pm$2.17  
&  69.76$\pm$2.26  
&  84.10$\pm$1.55  
&  85.07$\pm$1.15  
&  92.47$\pm$2.31
& 42.57$\pm$3.08
& 28.75$\pm$1.09
\\ 

\textbf{Improv.} (MLP/GNN)
& \tcr{6.38}/\tcr{2.81}
& \tcr{6.65}/\tcr{1.55}
& \tcr{2.25}/\tcr{1.61}
& \tcr{8.41}/\tcr{0.87}
& \tcr{10.70}/\tcr{0.34}
& \tcr{9.03}/\tcr{1.00}
& \tcr{2.81}/\tcr{1.05}
\\ \midrule

% \textbf{Improv.}  
% &  \tcr{15.20}     
% &  \tcr{12.45}     
% &  \tcr{3.35}      
% &  \tccy{-1.20}    
% &  \tcr{16.25}       \\ \midrule

$\textrm{MLP}_\textrm{GCNII}$      
&  62.41$\pm$2.24  
&  60.36$\pm$2.82  
&  78.91$\pm$1.34  
&  69.16$\pm$2.44  
&  80.69$\pm$2.11
& 36.62$\pm$2.61
& 25.47$\pm$1.26
\\

\bikit-$\textrm{MLP}_\textrm{GCNII}$      
&  66.91$\pm$3.38  
&  63.69$\pm$3.51  
&  84.92$\pm$0.62  
&  83.51$\pm$0.81 
&  88.71$\pm$2.25
& 40.92$\pm$2.35
& 30.56$\pm$0.76
\\ 

GCNII     
&  80.01$\pm$0.31  
&  67.39$\pm$1.19  
&  84.98$\pm$0.76  
&  83.41$\pm$1.25  
&  91.94$\pm$0.77
& 39.12$\pm$2.45
& 27.40$\pm$3.38
\\

\bikit-GCNII 
&  81.02$\pm$0.90  
&  68.22$\pm$1.60  
&  85.66$\pm$0.89  
&  84.53$\pm$2.15  
&  92.90$\pm$0.37
& 41.84$\pm$2.67
& 29.41$\pm$2.37
\\

\textbf{Improv.} (MLP/GNN)
&\tcr{4.50}/\tcr{1.01}
&\tcr{3.33}/\tcr{0.83}
&\tcr{6.01}/\tcr{0.68}
&\tcr{14.35}/\tcr{1.12}
&\tcr{8.02}/\tcr{0.96}
&\tcr{4.30}/\tcr{2.72}
&\tcr{5.09}/\tcr{2.01}
\\

\midrule

% \textbf{Improv.}  
% &  \tcr{15.01}     
% &  \tcr{10.65}     
% &  \tcr{5.55}      
% &  \tcr{4.74}      
% &  \tcr{16.84}       \\ \midrule

$\textrm{MLP}_\textrm{MixHop}$      
&  59.81$\pm$3.61  
&  55.60$\pm$3.88  
&  79.38$\pm$1.00  
&  59.56$\pm$9.23  
&  72.27$\pm$5.12       
& 26.81$\pm$3.47
& 22.37$\pm$1.24
\\

\bikit-$\textrm{MLP}_\textrm{MixHop}$      
&  65.47$\pm$3.81  
&  65.24$\pm$3.52  
&  84.22$\pm$0.53  
&  78.55$\pm$1.74  
&  87.45$\pm$1.74       
& 31.56$\pm$4.33
& 29.34$\pm$2.04
\\ 

MixHop     
&  71.27$\pm$1.87  
&  63.11$\pm$2.16  
&  82.42$\pm$0.27  
&  82.90$\pm$1.64  
&  89.55$\pm$1.50       
& 36.05$\pm$7.77
& 27.80$\pm$3.32
\\

\bikit-MixHop 
&  75.43$\pm$1.20  
&  66.28$\pm$1.69  
&  84.68$\pm$1.34  
&  85.58$\pm$1.26                 
&  90.44$\pm$0.93       
& 38.18$\pm$4.81
& 28.49$\pm$1.45
\\

\textbf{Improv.} (MLP/GNN)
&\tcr{5.66}/\tcr{4.16}
&\tcr{9.64}/\tcr{3.17}
&\tcr{4.84}/\tcr{2.26}
&\tcr{18.99}/\tcr{2.68}
&\tcr{15.18}/\tcr{0.89}
&\tcr{4.75}/\tcr{2.13}
&\tcr{6.97}/\tcr{0.69}
\\

\bottomrule

% \textbf{Improv.}  
% &  \tcr{13.57}     
% &  \tcr{12.20}     
% &  \tcr{4.85}      
% &  \tccy{-0.22}     
% &  \tcr{15.58}       \\ \bottomrule
\end{tabular}
\label{tab:inductive_main_reuslts}
\end{table*}

\noindent For proof of \textbf{Proposition}~\ref{thm::generalization}, please refer to the Appendix \ref{appendix::proof}. 

As we can see from \textbf{Proposition}~\ref{thm::generalization}, the generated distribution by $G$ could improve the generalization performance of the model trained in the source domain when applied to the target domain. It means the generator could facilitate the host GNN and $\MLPG$ to adapt to each other's distribution, i.e., achieving the knowledge transfer between the two models. 

\noindent\textbf{Complexity Analysis of BiKT.}
Since no additional computational units are introduced for the GNN and $\MLPG$, there is no extra
 computational complexity introduced in one training phase of the GNN and $\MLPG$. The main computational overhead of \bikit~comes from multiple iterative training. Due to the fact that the computational complexity of the GNN and MLP varies depending on the specific GNN used, let's denote the computational complexity of GNN to be $O(O_{gnn})$ and the computational complexity of MLP and generators to be $O(O_{mlp})$. Then the computational complexity of \bikit~during training can be roughly expressed as $O(tO_{mlp}+3tO_{gnn})$, where $t$ is the number of iterations.
 
Take GCN as an example, let $n$ denote the total number of nodes, $m$ be the total number of edges, and $L$ be the number of layers. For simplicity, the dimensions of the node hidden features remain constant as $d$. The complexity of GCN is $O(Lmd + Lnd^2)$, and the complexity of $\MLPG$ is $O(Lnd^2)$~\cite{wu2020comprehensive}. Then we have the complexity of BiKT as $O(tLmd + 3tLnd^2)$.
.

It should be noted that the computational complexity of the BiKT-enhanced GNN is equivalent to that of the original GNN during inference. If we directly use the derived MLP for inference, the computational complexity will be significantly less than the original GNN.
% whether the source model can benefit from the captured distribution of the generator.

\section{Experimental Results and Analysis}
\label{sect::experiment}
%In this section, we evaluate the performance of NFGNN on several real-world graph datasets. We first detail our experimental protocol, and then present the comparison results of NFGNN with the state-of-the-art methods. In addition, we perform various ablation studies to further verify the performance of proposed NFGNN.
%The experiments and models are performed on a worksation with an NVIDIA 2080Ti GPU and an Intel Xeon E5-2680 CPU.

\subsection{Experimental Settings}

\begin{table}
\centering
\caption{Results of the node classification task on the large-scale datasets.}
\label{tab::large_graph_reuslts}
\begin{tabular}{@{}ccccc@{}}
\toprule
Datasets               & Eval           & SAGE              & \bikit-SAGE        &  \textbf{Improv.} \\ \midrule
\multirow{2}{*}{OGB-Arxiv} & \textit{transductive}   &  74.55$\pm$0.69   &  75.47$\pm$0.21     &  $\uparrow$ \tcr{0.92}   \\
                       & \textit{inductive} &  71.37$\pm$0.71   &  71.71$\pm$0.46     &  $\uparrow$ \tcr{0.34} \\ \midrule 
\multirow{2}{*}{OGB-Products} & \textit{transductive}   &  78.98$\pm$0.14   &  79.69$\pm$0.27     &  $\uparrow$ \tcr{0.71}   \\
                       & \textit{inductive} &  76.98$\pm$0.41   &  77.47$\pm$0.41     &  $\uparrow$ \tcr{0.49} \\ 
                       \bottomrule
\end{tabular}
\vspace{-5pt}
\end{table}

\textbf{Datasets and Model Architectures.}
Seven widely used node classification benchmarks are adopted in our experiments, including three citation networks (Cora, Citeseer, and Pubmed~\cite{sen2008collective,yang2016revisiting}), two product co-occurrency networks (A-computer and A-photo~\cite{pitfalls}), two heterophilic networks (Chameleon and Squirrel), and two large-scale OGB datasets (OGB-Arxiv and OGB-Products~\cite{OGB}). 
% The details about the above datasets are provided in the supplemental material. 
Meanwhile, to validate the knowledge transfer capability of \bikit~under different architectures, several typical GNNs with different architectures are adopted, including GCN~\cite{GCN}, GAT~\cite{GAT}, FAGCN~\cite{FAGCN}, GCNII~\cite{GCNII}, and MixHop~\cite{mixhop}.
Besides, we also adopt two bioinformatics datasets and a large dataset from OGB~\cite{OGB}, including MUTAG, PTC, and OGB-molhiv, to verify the effectiveness of \bikit on the graph classification task. GCN and GIN as two classical GNNs for graph classification are employed as backbones. 

\noindent\textbf{Experimental Setup.}
To comprehensively evaluate our method, the dataset is split into training/validation/testing using the  \textbf{sparse split ratio (2.5\%/2.5\%/95\%)} for semi-supervised node classification.
Notably, for the large-scale Arxiv and Products datasets, we follow the same official splitting provided in OGB~\cite{OGB}.
For the experiments, we conduct the node classification task in two settings: transductive (\textit{tran}) and inductive (\textit{ind}) setting. Specifically, we follow the setting in~\cite{GLNN}, for the inductive setting, there are 80\% samples of the test set can be seen in the training phase, while the rest do not participate in the training. 
% Details of the two settings are supplied in \textbf{XXX}. 
We report the mean and standard deviation of 10 independent runs performed with different random seeds.
Accuracy is used to measure the model performance. For the graph classification task, following the experimental setting in~\cite{GIN}, we report the average and standard deviation of validation accuracies across the 10 folds within the cross-validation for the MUTAG and PTC datasets. For the OGB-molhiv dataset, we adopt the official splitting provided in OGB~\cite{OGB} and also report the mean and standard deviation of 10 independent runs. 

\noindent\textbf{Baseline Implementation.} For FAGCN, GCN, and GAT, we directly use the open-source codes released in~\cite{FAGCN}. For the others, we re-implement the models that refer to the open source code based on Deep Graph library~\cite{DGL}. For each model, we use hyperopt~\cite{hyperopt} to search for the optimal hyperparameters. Specifically, the search space for each hyperparameter is: learning rate within $\{0.001, 0.005, 0.01, 0.05\}$, dropout rate with $\{0.1, 0.2, 0.3, 0.4, 0.5, 0.6, 0.7, 0.8, 0.9\}$, weight decay rate within $\{1e-5, 5e-5, 1e-4, 5e-4, 1e-3, 5e-3\}$, hidden units within $\{16, 32, 64, 128, 256\}$.

% \usepackage{booktabs}

% \usepackage{booktabs}
% \usepackage{multirow}
\begin{table}[t]
\caption{Results of the graph classification task on three datasets.}
\label{table::GC_GNN}
\centering
\resizebox{2.5in}{!}{
\begin{tabular}{@{}c|ccc@{}}
\toprule
Method     & MUTAG & PTC & OGB-molhiv\\ \midrule
GCN
& 84.16 $\pm$ 6.40
& 61.86 $\pm$ 5.21
& 75.25 $\pm$ 2.07  \\

\bikit-GCN
& 87.26 $\pm$ 5.01
& 65.00 $\pm$ 4.67
& 77.15 $\pm$ 1.22  \\

\textbf{Improve.}
& \tcr{3.10}
& \tcr{3.14}
& \tcr{1.90} \\ \midrule

GIN
& 87.22 $\pm$ 7.45
& 62.31 $\pm$ 6.29
& 76.09 $\pm$ 1.43  \\

\bikit-GIN
& 90.13 $\pm$ 7.28
& 65.47 $\pm$ 5.99
& 76.47 $\pm$ 1.81  \\

\textbf{Improve.}
& \tcr{2.91}
& \tcr{3.16}
& \tcr{0.38} \\ \midrule
\end{tabular}}
\end{table}

\subsection{Performance Comparison}

For the node classification task, the performance in the transductive setting of our method on five popular benchmarks is presented in Table~\ref{tab:transductive_main_reuslts}.
It can be observed that, after knowledge transfer between the host GNN and the latent MLP, the \bikit-GNN under the five architectures consistently outperforms the original models on all datasets by a large margin, e.g., exceeding the MixHop by 6.86\% on the Citeseer dataset.
On the datasets with large-scale, we adopt GraphSAGE as the backbone and present the experimental results under two settings in Table~\ref{tab::large_graph_reuslts}.
It can be seen that the proposed \bikit~can still bring some improvement to the GNN on large-scale graphs.
% Moreover, MLPs are usually considered inferior to GNNs when handling graph data due to the lack of topology information.
Intriguingly, our \bikit-$\textrm{MLP}_\textrm{GNN}$ also achieves significant performance improvements on some datasets, even surpassing \bikit-GNN, e.g., achieving a 2.87\% improvement on the Citeseer dataset under the GCN architecture.
With the above promising results, it can be concluded that: 
1) \bikit~could be helpful for GNN to further leverage the capabilities of the \bfT operation to capture information from node content features without modifying the existing architecture of the GNN.
2) With the effective integration of topological information, MLPs could be as good as GNNs. This implies that further study of graph-based MLPs on other graph-related tasks is also a worthwhile research direction.

Meanwhile, we also evaluate the performance comparison of MLPs under the inductive setting to verify would the enhanced MLP via \bikit~perform well without the explicit topology guidance.
It can be seen from Table~\ref{tab:inductive_main_reuslts} that regardless of the GNN architecture, the \bikit-$\textrm{MLP}_\textrm{GNN}$ outperforms the MLP by a significant margin in most cases, e.g., obtaining an improvement in the range from \textbf{0.60\% to 10.30\%} on the five datasets under the GCNII architecture. In addition, \bikit-$\textrm{MLP}_\textrm{GNN}$ also shows its superiority compared to $\textrm{MLP}_\textrm{GNN}$.
The remarkable performance indicates that \bikit~can not only effectively transfer the topological knowledge from GNN to $\textrm{MLP}_\textrm{GNN}$, but also improve the ability of $\textrm{MLP}_\textrm{GNN}$ to capture knowledge from the node content features, enabling it to perform well even when there is no topological structure.

It can be seen from Table~\ref{table::GC_GNN} that \bikit also has the ability to enhance the performance of GNNs for the graph classification task. More concretely, GCN and GAT also achieved performance gains of 1.90\% and 0.38\% on OGB-molhiv, respectively. It indicates that the graph classification task can also benefit by strengthening the feature transformation operations of GNNs except for strengthening the feature propagation operations.

\begin{figure}
   \centering
   \subfigure[GCN-Cora]{
   \includegraphics[width=30mm]{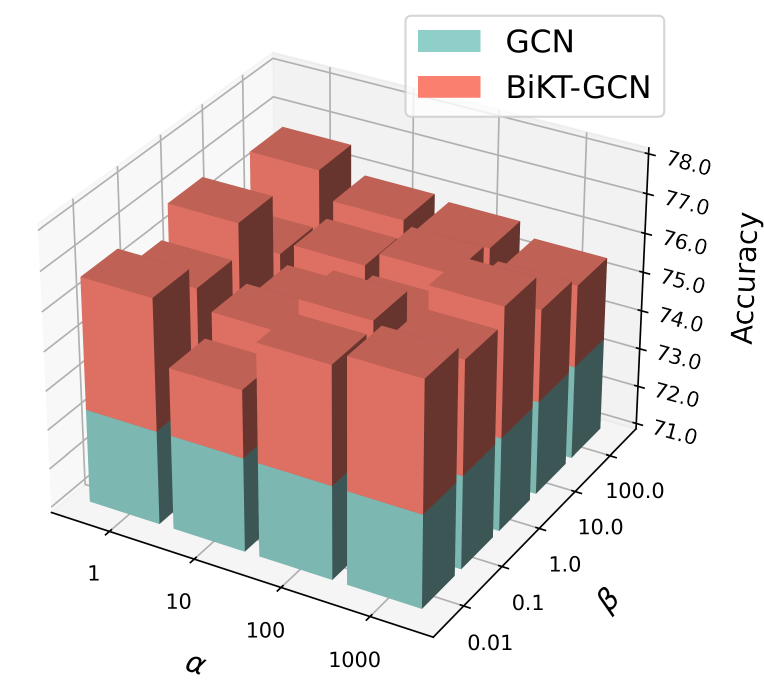}
   }
   % \hspace{-5mm}
   \subfigure[GCN-Citeseer]{
   \includegraphics[width=30mm]{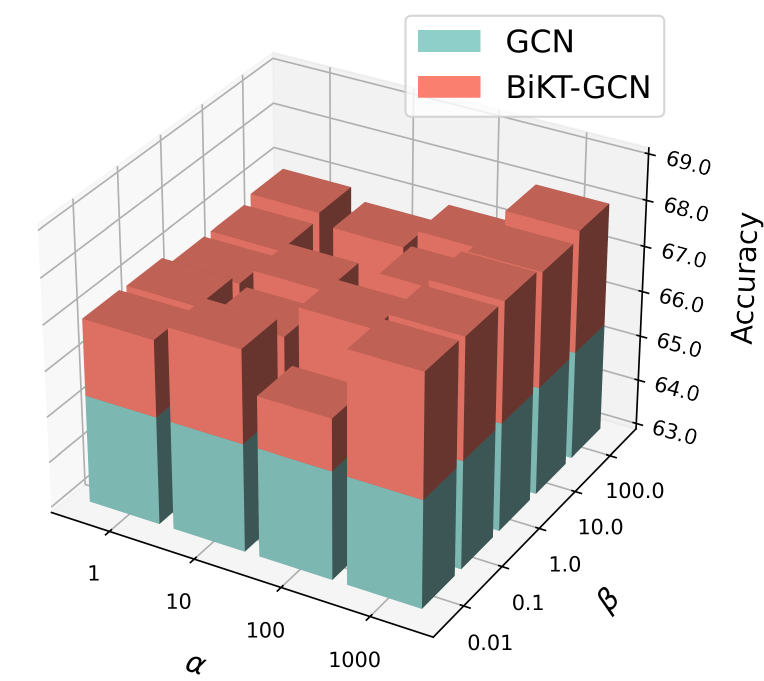}
   }
   \hspace{-5mm}
   \\
   \subfigure[MixHop-Cora]{
   \includegraphics[width=30mm]{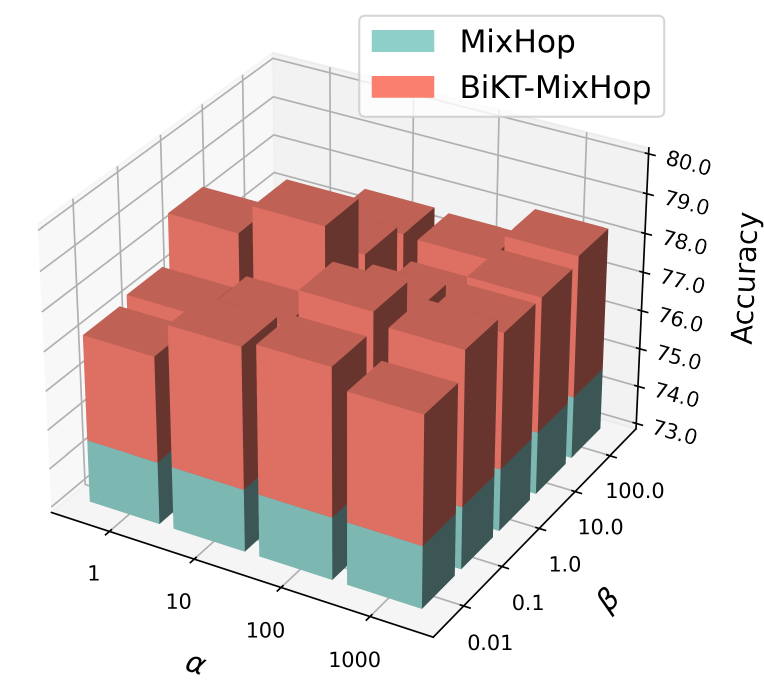}
   }
   % \hspace{-5mm}
   \subfigure[MixHop-Citeseer]{
   \includegraphics[width=30mm]{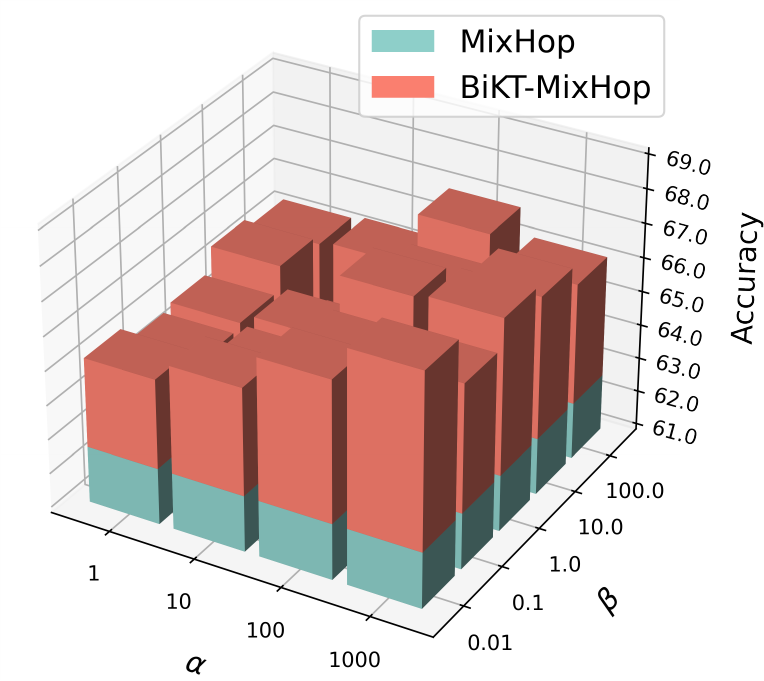}
   }
    \caption{The effect of different values of $\alpha$ and $\beta$ on performance gain.}
    
    \label{fig:parameter_analysis}
    % \vspace{-25pt}
\end{figure}

\subsection{Sensitivity Analysis of \bikit}
\textbf{Hyperparameter Analysis.} 
We analyze the impact of the coefficients $\alpha$ and $\beta$ on model performance using GCN and MixHop architectures.
As shown in Fig.~\ref{fig:parameter_analysis}, the \bikit-GCN holds a clear performance improvement over the original model, regardless of the parameter settings.
The inspiring results show that the generalization ability of our \bikit-GCN is not heavily reliant on hyperparameter tuning. 
More concretely, both GCN and Mixhop behave relatively sensitive to $\alpha$ on the Citeseer dataset, and the larger the $\alpha$ is, the greater the performance gain. While the two GNNs seem to be more impacted by $\beta$ on the Cora dataset.
\begin{figure}[t]
\subfigure[GCN]{
   \includegraphics[width=28mm]{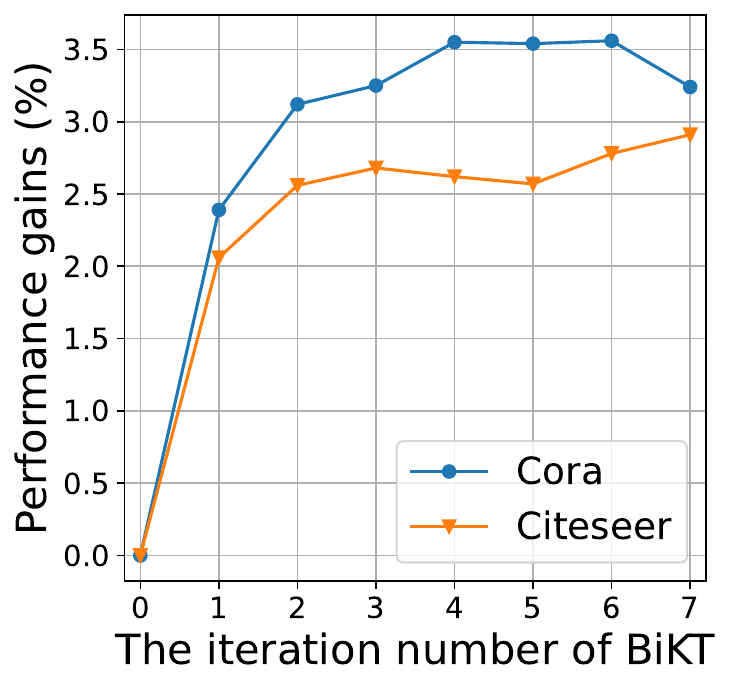}
   }
   \hspace{-5mm}
   \subfigure[GAT]{
   \includegraphics[width=28mm]{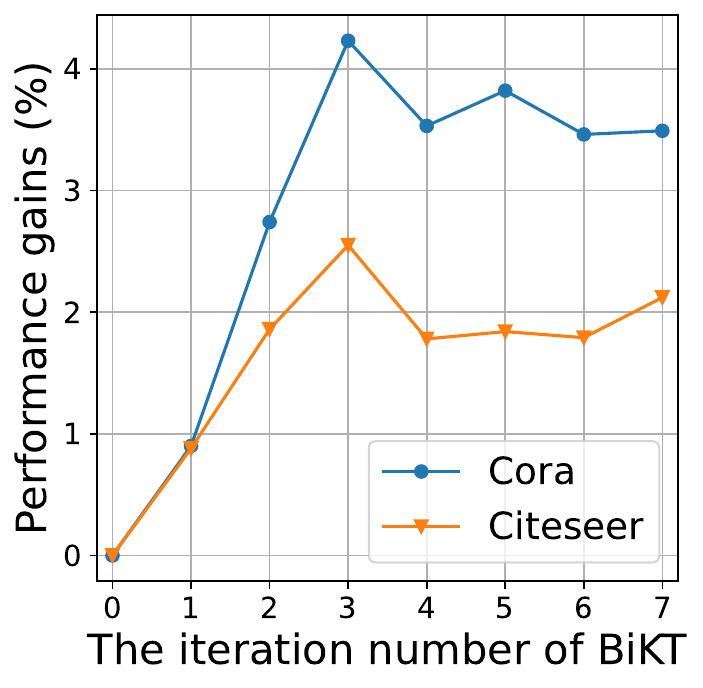}
   }
   \hspace{-5mm}
   \subfigure[MixHop]{
   \includegraphics[width=28mm]{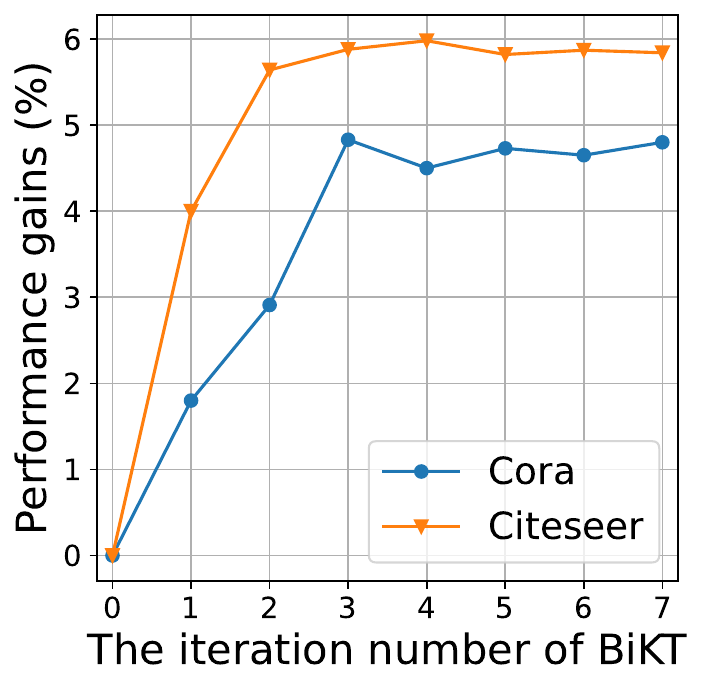}
   }
   % \hspace{-5mm}
   % \captionstyle{normal}
    \caption{The performance gains when performing the number of iterations in the recurrent training from 0 to 7.}
    \label{fig::Iteration_number}
\end{figure}

\begin{figure}[t]
\vspace{-10pt}
   \subfigure[GCN]{
   \includegraphics[width=30mm]{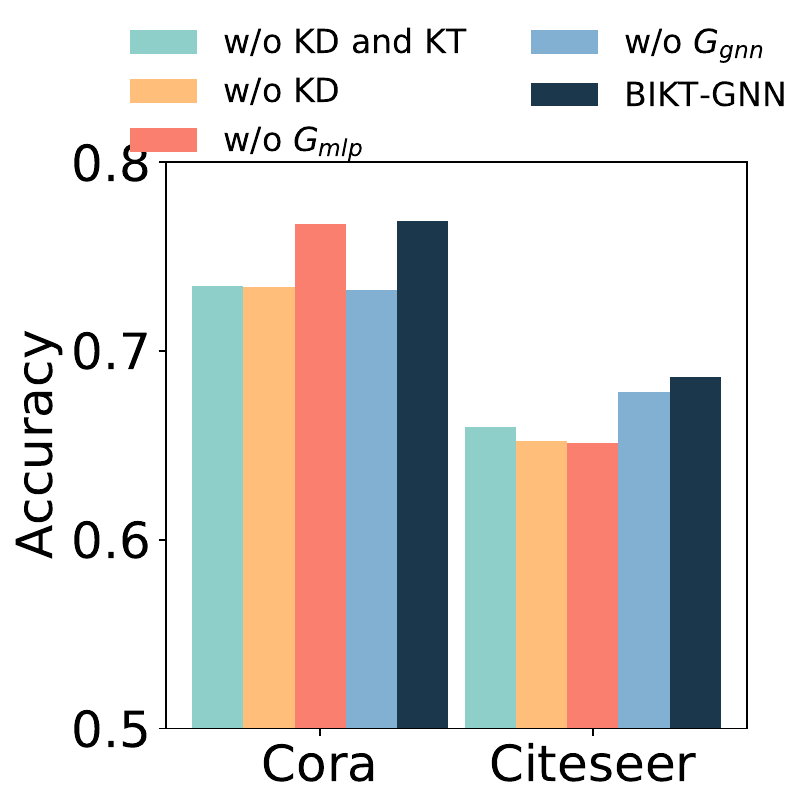}
   }
   \hspace{-5mm}
   \subfigure[GAT]{
   \includegraphics[width=30mm]{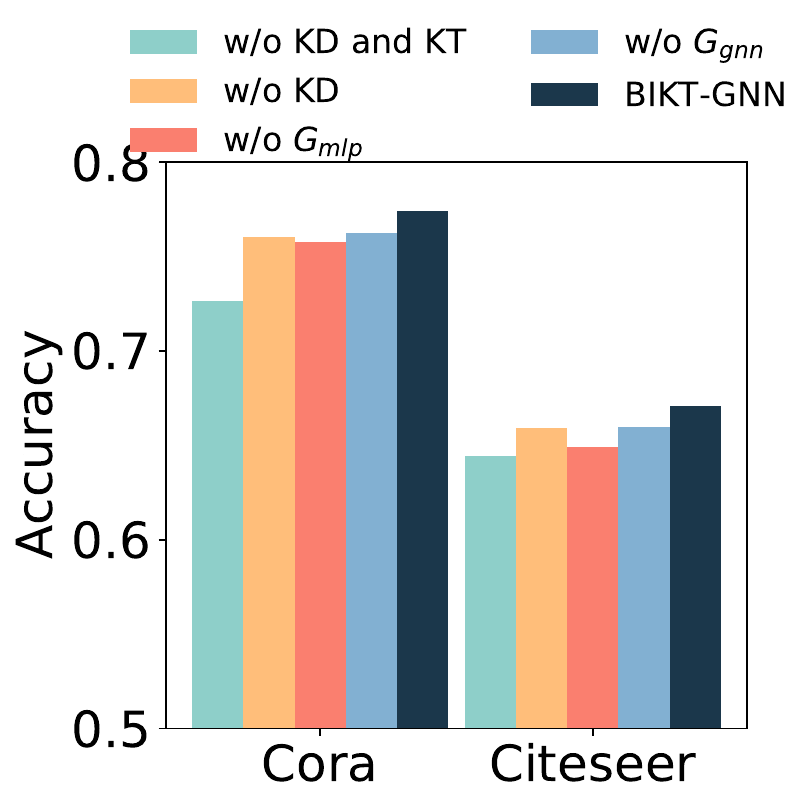}
   }
   \hspace{-5mm}
   \subfigure[MixHop]{
   \includegraphics[width=30mm]{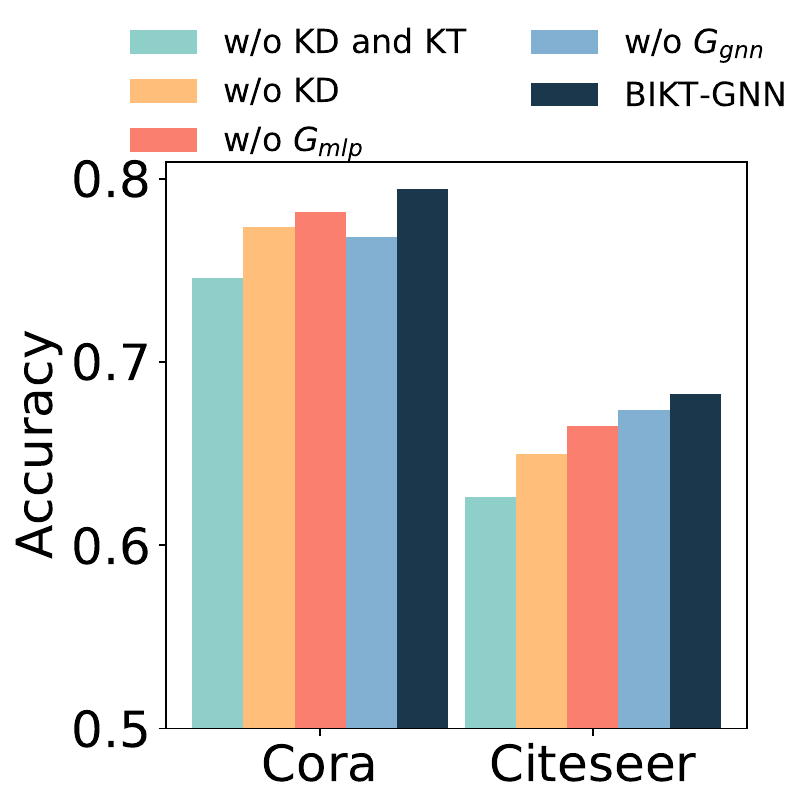}
   }
\caption{ Effectiveness evaluation of knowledge transfer (KT) and knowledge distillation (KD), where knowledge transfer denotes specifically the process of representation generation via generators $G_{gnn}$ and $G_{mlp}$. }
\label{fig::Ablation}
\end{figure}

% \begin{table}[t]
% \caption{Results of the graph classification task, where \tcr{red} letters denote the performance gains compared to the base GNN, respectively.}
% \label{table::GC_GNN}
% \centering
% \begin{tabular}{@{}c|ccc@{}}
% % \toprule
% % Method     & GCN & \bikit-GCN & Improve.\\ \midrule
% % MUTAG 
% % & 84.16 $\pm$ \tiny{6.40}
% % & 87.26 $\pm$ \tiny{5.01}
% % & $\uparrow$ \tcr{3.10}                 \\

% % PTC  
% % & 61.86 $\pm$ \tiny{5.21}
% % & 65.00 $\pm$ \tiny{4.67}
% % & $\uparrow$ \tcr{3.14}  \\

% % OGB-molhiv
% % & 75.25 $\pm$ \tiny{2.07}
% % & 77.15 $\pm$ \tiny{1.22}
% % & $\uparrow$ \tcr{1.90} \\ \midrule

% \end{tabular}
% \end{table}

\noindent\textbf{Analysis of the Recurrent Training.} We report the performance gains when performing the number of iterations from 0 to 7 to verify the role of the recurrent training strategy for \bikit. We can see from Fig.~\ref{fig::Iteration_number} that the performance gains of GCN, GAT, and MixHop on both datasets stabilize as the number of iterations increases. As post a certain extent of knowledge transference, the latent potential of the feature transformation operations will be thoroughly harnessed, thus stabilizing the performance.

\begin{table}[t]
\caption{The MMD distances between the representation distribution of generators and models at different training epochs.}
\centering
\label{tab::MMD_gen}
\resizebox{3in}{!}{
\begin{tabular}{@{}clccc@{}}
\toprule
                                  & Dataset & Initial     & 50-th Epoch    & 100-th Epoch   \\ \midrule
\multirow{2}{*}{MMD(GCN, $G_{gnn}$)} & Cora                        & 4.521 & 1.531 & 0.746 \\
                                  & Citeseer                    & 5.079 & 1.262 & 0.530 \\ \midrule
\multirow{2}{*}{MMD(MLP, $G_{mlp}$)} & Cora                        & 4.435 & 0.642 & -     \\
                                  & Citeseer                    & 4.615 & 1.527 & 0.532 \\ \midrule
\multirow{2}{*}{MMD(GAT, $G_{gnn}$)} & Cora                        & 4.623 & 1.270 & 0.638 \\
                                  & Citeseer                    & 6.339 & 1.262 & -     \\ \midrule
\multirow{2}{*}{MMD(MLP, $G_{mlp}$)} & Cora                        & 4.800 & 2.513 & 0.955 \\
                                  & Citeseer                    & 5.127 & 0.960 & -     \\ 
                                  \bottomrule 
\end{tabular}}
\end{table}

% \begin{table}[t]
% \caption{The MMD distances between the representation distribution of GNN and $\MLPG$ after multiple iterations.}
% \centering
% \begin{tabular}{@{}clccc@{}}
% \toprule
%                                & \multicolumn{1}{c}{Dataset} & Initial     & 1-st Iteration     & 2-nd Iteration     \\ \midrule
% \multirow{2}{*}{MMD(GCN, MLP)} & cora                        & 2.529 & 0.972 & 0.942 \\
%                                & citeseer                    & 4.096 & 2.730 & 1.467 \\ \midrule
% \multirow{2}{*}{MMD(GAT, MLP)} & cora                        & 3.707 & 2.764 & 0.974 \\
%                                & citeseer                    & 4.344 & 2.221 & 1.711 \\ \bottomrule
% \end{tabular}
% \end{table}

\subsection{Ablation Study}
\noindent\textbf{Effectiveness of knowledge extraction.}
To analyze the contributions of knowledge transfer (KT) and knowledge distillation (KD) in Eq.~\eqref{obj::loss_MLP}, an ablation study of \bikit~is conducted on Cora and Citeseer datasets. As shown in Fig.~\ref{fig::Ablation}, each term of Eq.~\eqref{obj::loss_MLP} has a positive effect on GNN. Interestingly, for GCN as the base GNN, either part alone has a general gain on two datasets, but a more significant gain is achieved when both are used together. It can also be observed that, the impact of $G_{gnn}$ is stable and moderate, regardless of the GNN used. On the contrary, the impact of $G_{mlp}$ varies widely with architecture and dataset.
% \footnote{In addition to the above experiments, we provide further experimental details in the supplementary material, as well as experiments of node classification on the heterophilic graph and graph classification.}
% It verifies empirically that GNN can benefit from the captured distribution according to \bikit.

\begin{figure}
\centering
  \subfigure[GCN]{
   \includegraphics[width=26mm]{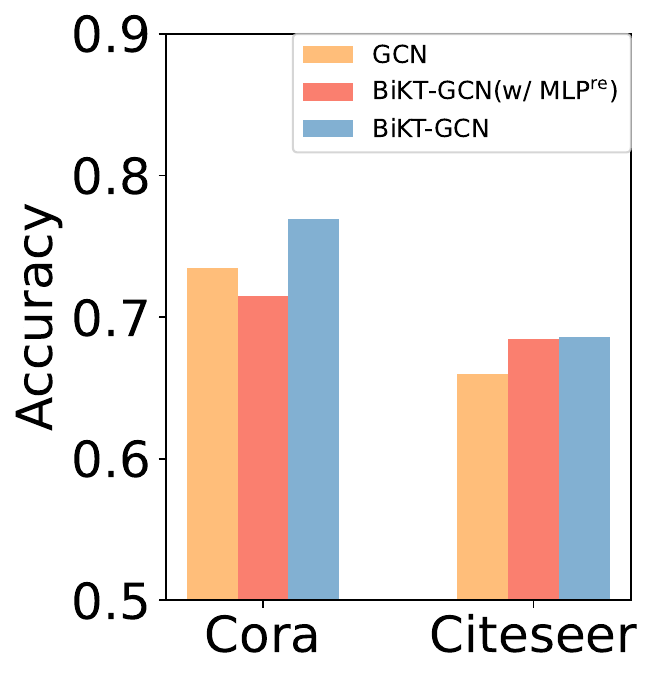}
   }
   \hspace{-5mm}
   \subfigure[GAT]{
   \includegraphics[width=26mm]{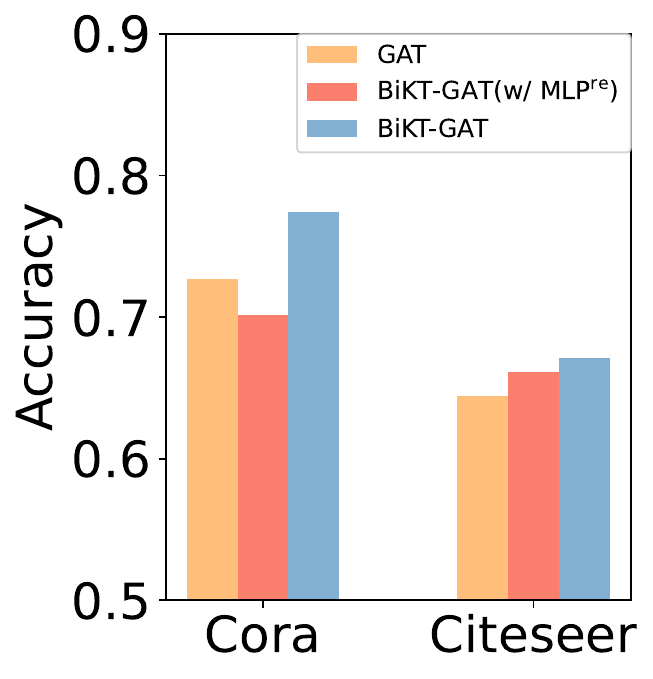}
   }
   \hspace{-5mm}
   \subfigure[MixHop]{
   \includegraphics[width=26mm]{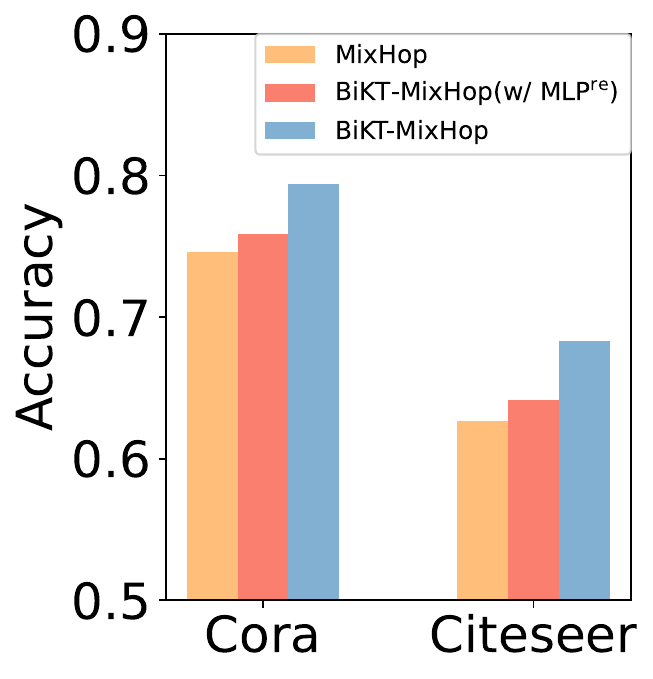}
   }
    \caption{The effectiveness evaluation of parameter inheritance stagy.}
    \label{fig::parameter}
\end{figure}

\noindent\textbf{Effectiveness of parameter inheritance.} 
To illustrate the necessity of sharing parameters between the GNN and $\MLPG$, we perform a comparison of GNN with re-initialized MLP (denoted as BiKT-GNN w/ $\textrm{MLP}^{re}$) in the BiKT framework. As shown in Fig.~\ref{fig::parameter}, the performance gains brought by BiKT-GNN w/ $\textrm{MLP}^{re}$ are modest compared to BiKT-GNN. Besides, it may instead degrade the model's performance in some cases. The experimental results illustrate that the parameter inheritance scheme has a more stable performance, in comparison to re-initialization. 

\subsection{Analysis of Generators for Distribution Modeling}
To quantitatively analyze whether the generator is capable of efficiently modeling the representation distribution of the model with Eq.~\eqref{obj::loss_G}, we calculate the MMD distances between the representation generated by generators at different training epochs and the representation output by the model. The results are reported in Table \ref{tab::MMD_gen}. It can be observed that as the training of the generator progresses, the distribution of its generated representations gradually becomes closer to the target representation distribution. This indicates the effectiveness of the optimization objective set for the generator and aligns with our expectations. 

\section{Discussion}
% \textbf{Connection to Previous Studies.} 
% Intuitively, \bikit~can also be seen as a data augmentation-based approach since the generative model is used to learn the representation distribution for knowledge transfer.
Recently, a lot of efforts have been made in data augmentation over graphs~\cite{zhao2021data,feng2020graph,verma2021graphmix}.
In general, existing data augmentation methods start from the topological structure of the graph and propose a series of methods to perturb the connectivity of the graph~\cite{rong2019dropedge,chen2020measuring,zhao2021data}.
Differently, our \bikit~conducts the augmentation from the representation space without modifying the topological structure and node features. \bikit~is therefore orthogonal to these methods and can be used simultaneously. 
Particularly, similar ideas have also been applied to other fields, such as image translation~\cite{hoffman2018cycada} and federal learning~\cite{FedGen}.
% several studies in image-to-image translation~\cite{hoffman2018cycada,10050808,lin2023cycle} also proposed the above similar suggestion. 

In addition, several works have also emerged in recent years aimed at exploring the connection between GNNs and MLPs~\cite{PMLP,MLPInit} and how MLPs can be used as an alternative to GNNs on graph-related tasks~\cite{GAMLP,GLNN,chen2020graph}. To this end, these approaches introduce techniques such as regularization~\cite{graph-mlp} and data augmentation~\cite{yourself} to improve the performance of MLP on graph data. 
% Differently, we aim to find a way in this paper to make GNNs and the derived model therein complement and promote each other, and propose to understand the connection between GNNs and MLPs from a perspective of domain adaption. Particularly, ~\cite{MLPInit} can be seen as a special case of \bikit.
In contrast, the objective of this paper is to discover a method to enhance GNNs through feature transformation and $\MLPG$. We also propose to examine the relationship between GNNs and MLPs through the perspective of domain adaptation. Notably, the approach presented in ~\cite{MLPInit} can be viewed as a specific instance of \bikit.

% \textbf{Limitation.} As a framework designed to enhance the GNN family in general, it should be noted that \bikit~is not completely model-agnostic. Concretely, \bikit~requires GNN to have a classifier that is placed after \bfP operations. It means that APPNP-like GNNs~\cite{APPNP}, such as GPRGNN~\cite{GPRGNN} and DAGNN~\cite{DAGNN}, may have difficulty benefiting from \bikit~unless we add an additional classifier to it as was done in \cite{FAGCN}. Moreover, although \bikit~is very instrumental to GNN and the MLP therein, it has not yet reached the upper bound of performance in Fig.~\ref{fig::motivation_2}. This shows that there is still scope for progress in the knowledge transfer between GNNs and the inherent \bfT operations.
% i.e., the accuracy of union $\mathrm{GNN} \bigcup \mathrm{MLP}_\mathrm{GNN}$ shown in Fig~\ref{fig::motivation_2}.

\section{Conclusion}
In this work, we point out the existing GNNs have not effectively unleashed the potential of feature transformation operations therein with the empirical investigation. To address this issue, we propose a generic framework, \bikit, to capture the induced distribution from the GNN and the derived model composed by feature transformation operations, thus improving them together. Moreover, we introduce a new perspective from domain adaption to unpack the connection between GNNs and MLPs, and provide a theoretical analysis of our approach. 
Extensive experiments on 7 datasets with 6 GNNs as backbones demonstrate that \bikit~is not only highly conducive for GNN to further leverage the capabilities of the \bfT operation, but also greatly boost the performance of latent MLP derived from the GNN. 

%\textcolor{black}{Studying spectral-based GNNs in accordance with the idea of graph signal processing theory is one of the origins of GNNs. With different starting points, the spatial-designed GNNs aim to design the neighborhood aggregation mechanism based on the topological characteristics of the graph, focusing more on the local relationship between nodes and their neighbors. In contrast, spectral-based GNNs are dedicated to the design of the filtering of the graph signal in the spectral domain, analyzing the graph more from a global perspective.} 
% 

% The proposed NFGNN in this paper provides a new form of trade-off between global and local perspectives in the spectral domain. Particularly, NFGNN can be seen as an extension of the existing methods for estimating global filters. 
%Some studies~\cite{GWNN,framelets} achieve similar objectives via graph wavelets, but in comparison, our approach is more concise and flexible.

% \ifCLASSOPTIONcompsoc
%   % The Computer Society usually uses the plural form
%   \section*{Acknowledgments}
% \else
%   % regular IEEE prefers the singular form
% \fi

% The authors would like to thank the reviewers for their comments
% and suggestions.

\ifCLASSOPTIONcaptionsoff
  \newpage
\fi



%\bibliographystyle{IEEEtran}
{
	\bibliographystyle{IEEEtran}
	\bibliography{neurips_2021.bib}
}

% \clearpage

\appendix
\label{appendix::proof}
\subsection*{Derivations of Proposition 4.1}

\textbf{Proposition 4.1.} \textit{ Let $q(\rvz|y)$ be the distribution modeled by generator $G$, $p(y|\rvz;f_{cls})$ and $p(\rvz|y;f_{cls})$ be the posterior distribution and the corresponding induce distribution of $f_{cls}$. Then maximizing $H(q(\rvz|y))+\E_{y \sim p(y)} \E_{\rvz \sim q(\rvz|y)}[\log p(y|\rvz)]$ is equivalent to minimizing the conditional KL-divergence between $q(\rvz|y)$ and $p(\rvz|y;f_{cls})$, i.e., $\KL[q(\rvz|y)||p(\rvz|y;f_{cls})]$.}

\begin{proof}[\textbf{Proof}]
According to the definition of $\KL$, $\KL(q(\rvz|y)\Vert p(\rvz|y;f_{cls}))$ can be expanded as:
\begin{equation}
    \begin{aligned}
    \label{eq::KL_raw}
        & {\KL}[ q(\rvz|y) \Vert p(\rvz|y;f_{cls})] \\
        & = \E_{y \sim p(y)} \left[ \E_{\rvz \sim q(\rvz|y)} \left[ \log \frac{ q(\rvz|y)}{ p(\rvz|y;f_{cls})} \right] \right] \\
        & = \E_{y \sim p(y)} \E_{z \sim q(\rvz|y)} \left[ \log q(\rvz|y) \right ] \\
        &\ \ \ \ - \E_{y \sim p(y)} \E_{z \sim q(\rvz|y)} \left[   { \log p(\rvz|y;f_{cls})} \right]\\
        & = - H(q(\rvz|y)) - \E_{y \sim p(y)} \E_{z\sim q(\rvz|y)} [ \log p(\rvz|y;f_{cls})]
    \end{aligned}
\end{equation}
For the second term, it can be rewritten based on the Bayes Rule:
\begin{equation}
    \begin{aligned}
    \label{eq::KL_second_term}
&\E_{y \sim p(y)} \E_{z\sim q(\rvz|y)} [ \log p(\rvz|y;f_{cls})] \\
& = \E_{z\sim q(\rvz|y)} [ \log \frac{p(y|\rvz;f_{cls})  p(\rvz)}{p(y)}   ] \\
& = \E_{y \sim p(y)} \E_{z\sim q(\rvz|y)} [ \log p(y|\rvz;f_{cls}) + {\log p(\rvz)- \log p(y)}]
    \end{aligned}
\end{equation}
Since $\log p(\rvz)$ and $ \log p(y)$ are the prior distribution that is constant w.r.t $q(\rvz|y)$, therefore, we can obtain the follows by substituting Eq.~\eqref{eq::KL_second_term} into Eq.~\eqref{eq::KL_raw}:
\begin{equation}
    \begin{aligned}
    \label{eq::KL_final}
    & \argmin_{q(\rvz|y)}{\KL}[ q(\rvz|y) \Vert p(\rvz|y;f_{cls})] \\
    & \equiv \argmin_{q(\rvz|y)} - H(q(\rvz|y)) \\
    &\ \ \ \ - \E_{y \sim p(y)} \E_{z\sim q(\rvz|y)} [ \log p(y|\rvz;f_{cls}) + {\log p(\rvz)- \log p(y)}] \\
    & \equiv \argmin_{q(\rvz|y)} - H(q(\rvz|y)) - \E_{y \sim p(y)} \E_{z\sim q(\rvz|y)} [ \log p(y|\rvz;f_{cls})] \\
    & \equiv \argmax_{q(\rvz|y)} H(q(\rvz|y)) + \E_{y \sim p(y)} \E_{z\sim q(\rvz|y)} [ \log p(y|\rvz;f_{cls})] \\
    \end{aligned}
\end{equation}
This completes the proof.
\end{proof}

\subsection*{Derivations of Proposition 4.2}
Let's first introduce some symbols and their definitions from~\cite{DA} as below:
\begin{table}[h]
\centering
     % \caption{Major symbols and definitions.} %
     \label{tab:dfn}
    \begin{tabular}{ lp{5cm}}
         \toprule
         \textbf{Symbols} &  \textbf{Definitions}\\
         \midrule
         $\gX$ & the raw feature space \\
         $\gZ$ & the representation space \\
         $\gY$ & the set of class labels \\
         $\gH$ & a set of hypothesis $h$ \\
         % $\gS$ & source domain with distribution $\gD_s$\\
         % $\gT$ & target domain with distribution $\gD_t$\\
         $\gD_G$ & an auxiliary distribution derived from the generator $G$ \\
         $B$ & a probability event \\
         $\gR$ & a fixed representation function from $\gX$ to $\gZ$ \\
         \midrule
         $d_{\gH}(\gD_1, \gD_2)$ & the $\gH$-divergence between $\gD_1$ and $\gD_2$ \\
         $\epsilon_{*}(h)$ & the expected risk of $h$ on the domain $*$ \\
         $\lambda = \underset{h \in \gH}{\min} \left( \epsilon_{\gT}(h) + \epsilon_{\gS}(h) \right )$ & the optimal risk on $\gT$ and $\gS$ \\
         \bottomrule
    \end{tabular}
    \end{table}

\noindent\textbf{Assumption 4.1.}
\textit{  Given $f_{cls}^{tgt}$ and $f_{cls}^{src}$ of the target and source models respectively, $q(\rvz|y)$ can approximate $p(\rvz|y;f_{cls}^{tag})$ and have $d_{\gH} (q(\rvz|y), p(\rvz|y;f_{cls}^{tag}))  \leq d_{\gH} (p(\rvz|y;f_{cls}^{src}), p(\rvz|y;f_{cls}^{tag}))$, where $d_{\gH}$ denotes the $\gH$-divergence.}

\noindent\textbf{Proposition 4.2.}
\textit{Suppose assumption 1 holds. Let $\gT$ and $\gS$ be the source domain and target domain with the distribution $\gD_{s}$ and $\gD_{t}$, respectively. Let $\gR$ be a representation function from $\gX$ to $\gZ$. 
Denote $\gD_{G}$ be an auxiliary distribution derived from a generator $G$ 
and $\gD_{s}^{'} = \tau \gD_{G} + (1-\tau) \gD_{s}$.
Denote $\gH$ be a set of hypothesis with VC-dimension $d$. Given an empirical dataset $\hat{\gD}_{s}$ and an augmented dataset $\hat{\gD}_{s}^{'}$ with $|\hat{\gD}_{s}|=m$ and $|\hat{\gD}_{s}^{'}|=m' > m$. 
Let $J(m,d) = \sqrt{\frac{4}{m} \left( d \log\frac{2 e m }{d} + \log \frac{4}{\delta} \right) }$, where $e$ is the base of the natural logarithm.
If $(\epsilon_{\gS}(h) - \epsilon_{\gS^{'}}(h))$ is bounded and $m > \frac{1}{2}d$, then with probability at least $1-\delta$, for every hypothesis $h \in \gH$:
\begin{equation}
    \begin{aligned} 
    % \epsilon_{\gT}(h) 
    % &\leq \varepsilon_{\gS'}(h) 
    % + \sqrt{\frac{4}{m'} \left( d \log\frac{2 e m' }{d} + \log \frac{4}{\delta} \right) }  
    %  %
    %  + d_{\gH} (\Dz_{s}^{'}, \Dz_{t}) + \lambda' \\
    %  & \leq \epsilon_{\gS}(h)  + \sqrt{\frac{4}{m} \left( d \log\frac{2 e m }{d} + \log \frac{4}{\delta} \right) } + d_{\gH} (\Dz_{S}, \Dz_{T}) + \lambda
    %%%%%%%%%%%%%%%%%%%%%%%%%%%%%%%%%%%%%%%%%%%%%%%%%%%%%%%%%%%%%%%%%%%%%
    \epsilon_{\gT}(h) 
    & \leq \hat{\epsilon}_{\gS'}(h) 
    + J(m',d)
     + d_{\gH} (\Dz_{s}^{'}, \Dz_{t}) + \lambda' \\
     & \leq \hat{\epsilon}_{\gS}(h)  
     + J(m,d)
     + d_{\gH} (\Dz_{s}, \Dz_{t}) + \lambda
    \end{aligned}
\end{equation}
where $\epsilon_{*}(h)$ and $\hat{\epsilon}_{*}(h)$ are the expected and empirical risk of $h$ on the domain respectively. $\gS'$ denotes the augmented source domain with $\Dz_{s}^{'}$, $\lambda = \underset{h}{\min} \left( \epsilon_{\gT}(h) + \epsilon_{\gS}(h) \right )$ and $\lambda' = \underset{h}{\min} \left( \epsilon_{\gT}(h) + \epsilon_{\gS'}(h) \right )$ is the optimal risk on two domains.}

\begin{proof}[\textbf{Proof}]
For the proof of Proposition 4.2, we first introduce Lemma~\ref{lemma:da-bounds} from~\cite{DA,DA2} to give the upper bound for the generalization performance of domain adaption (DA):
\begin{lemma}  \label{lemma:da-bounds}
    \textnormal{\textbf{Generalization Bounds for DA \cite{DA}:}}
    \\
    Let $\gH$ be a hypothesis space of VC-dimension $d$.
    Let $\gT_S$ and $\gT_T$ be the source and target domains, whose data distributions are $\gD_s$ and $\gD_t$.
    and $\Dz_s, \Dz_t$ be the induced images of $\gD_s$ and $\gD_t$  over $\gR$, respectively, s.t., $\E_{z \sim \hat{D}_{*}}[B(z)]=\E_{x\sim \gD_{*}}[B(\gR(x))]$ when given a probability event $B$, and so for $\Dz$. Given an observable dataset with $m$ samples,
    then with probability at least $1 - \delta$, $\forall~h \in \gH$:
    \begin{equation}
    \begin{aligned} 
    \label{lemma::raw}
    \epsilon_{\gT}(h) \leq \hat{\epsilon}_{\gS}(h) %
     & + \sqrt{\frac{4}{m} \left( d \log\frac{2 e m }{d} 
     + \log \frac{4}{\delta} \right) }  \\
     & + d_{\gH} (\Dz_{s}, \Dz_{t}) + \lambda,
    \end{aligned}
    \end{equation}
    %}
where $e$ is the base of the natural logarithm.
\end{lemma}

Through substituting $\gD_s^{'}$ and $\Dz_{s}^{'}$ into Eq.~\eqref{lemma::raw}, it is not hard to obatin:
\begin{align}
\label{eq::generated_D}
    \epsilon_{\gT}(h) 
    &\leq \hat{\epsilon}_{\gS'}(h) 
    + J(m',d)
     + d_{\gH} (\Dz_{s}^{'}, \Dz_{t}) + \lambda' 
\end{align}
The theorem will be proved if we can show that $J(m',d) < j(m,d)$ and $d_{\gH}(\Dz'_{s}, \Dz_t) < d_{\gH}(\Dz_{s}, \Dz_t)$ when assumption 1 holds with $(\epsilon_{\gS}(h) - \epsilon_{\gS^{'}}(h)) > 0$ is bounded and $m > \frac{1}{2}d$.

For $J^{2}(m,d) = \frac{4}{m} \left( d \log\frac{2 e m }{d} + \log \frac{4}{\delta} \right)$, it can be divided into two terms $\frac{4}{m} d \log\frac{2 e m }{d}$ and $\frac{4}{m}\log \frac{4}{\delta}$. For the first term,
let $x = \frac{m}{d}$, we have $\frac{4}{m} d \log\frac{2 e m }{d}  = \frac{4}{x} \log 2 e x$.
We can obtain the derivative of $\frac{4}{x} \log 2 e x$ with respect to $x$ as:
\begin{equation}
\label{eq::J(md)}
\begin{aligned}
\frac{\mathrm{d} {(\frac{4}{x} \log 2 e x})}{\mathrm{d} x} = -4x^{-2}\log2x
\end{aligned}
\end{equation}
From Eq.~\eqref{eq::J(md)} we can see that $\frac{4}{x} \log 2 e x$ is monotonically decreasing when $x > \frac{1}{2}$. Meanwhile, it is easy to find for the second term that $\frac{4}{m}\log \frac{4}{\delta}$ is also monotonically decreasing when $m > 0$. From the above we can find that $J(m',d) < J(m,d)$ if $m>\frac{1}{2}d$.
Moreover, according to the definition of $d_{\gH}(\cdot, \cdot)$, it can be derived for $d_{\gH} (\Dz_{s}^{'}, \Dz_{t})$ that:
\begin{equation}
\label{eq::d_{H}_2}
\begin{aligned}
    & d_{\gH} (\Dz_s', \Dz_t) \\
    & = 2 \sup_{\gA \in \gA_\gH} \left \vert \E_{\rvz \sim \Dz_s'}\left[\rp(\gA(\rvz) ) \right] - \E_{\rvz \sim \Dz_t} \left[\rp(\gA(\rvz) ) \right] \right \vert \\
    % \\
    % =  & 2 \sup_{\gA \in \gA_\gH}  \left \vert \E_{z \sim \tau \gD_{G} + (1-\tau) \gD_{s}}\left[ \rp(\gA(z)) \right] - \E_{z \sim \Dz_t} \left[\rp(\gA(z) ) \right] \right \vert \\
     % = & 2 \sup_{\gA \in \gA_\gH}  \left \vert \tau\E_{z \sim \Dz_s }\left[ \rp(\gA(z)) \right]  + (1-\tau)\E_{z \sim \Dz_G }\left[ \rp(\gA(z)) \right]  - \E_{z \sim \Dz_t} \left[\rp(\gA(z) ) \right] \right \vert \\
%      % %
%      % \leq &  \sup_{\gA \in \gA_\gH}  \left \vert  \E_{z \sim \Dz_s }\left[ \rp(\gA(z)) \right]   - \E_{z \sim \Dz} \left[\rp(\gA(z) ) \right] \right \vert  +  \sup_{\gA \in \gA_\gH}  \left \vert  \E_{z \sim \Dz_A }\left[ \rp(\gA(z)) \right]  - \E_{z \sim \Dz} \left[\rp(\gA(z) ) \right] \right \vert \\  
%      % = & \frac{1}{2} d_{\gH} (\Dz_s,\Dz) + \frac{1}{2} d_{\gH}(\Dz_A, \Dz).
% \end{aligned}
% \end{equation}
% % Since $\gD'_{s} = \tau \gD_{G} + (1-\tau) \gD_{s}$, Eq.~\eqref{eq::d_{H}_1} can be rewritten as:
% \begin{equation}
% \label{eq::d_{H}_2}
% \begin{aligned}
% a
% d_{\gH} (\Dz_s', \Dz_t) 
    &= 2 \sup_{\gA \in \gA_\gH}  | \tau\E_{\rvz \sim \Dz_s }\left[ \rp(\gA(\rvz)) \right] - \tau\E_{\rvz \sim \Dz_t} \left[\rp(\gA(\rvz) ) \right]  \\
    &\ \ \ \ + (1-\tau)\E_{\rvz \sim \Dz_G }\left[ \rp(\gA(\rvz)) \right]  - (1-\tau)\E_{\rvz \sim \Dz_t} \left[\rp(\gA(\rvz) ) \right] | \\
    & \leq  \tau d_{\gH} (\Dz_s,\Dz_t) + (1-\tau) d_{\gH}(\Dz_G, \Dz_t)\\ 
\end{aligned}
\end{equation}
According to Assumption 1 that $d_{\gH} (q(\rvz|y), p(\rvz|y;f_{cls}^{tag}))  \leq d_{\gH} (p(\rvz|y;f_{cls}^{src}), p(\rvz|y;f_{cls}^{tag}))$ and the definition of induce image $\Dz$ of $\gD$ over $\gR$, we have:
% , we can derive that:
% \begin{equation}
% \label{eq::KL4DA}
% \begin{aligned}
% % a
% & \KL[q(\rvz|y)||p(\rvz|y;f_{cls})] \\
% & = \KL[\sum_{y \in \gY}p(y)q(\rvz|y)||\sum_{y \in \gY}p(y)p(\rvz|y;f_{cls})] \\
% & = \KL[q(\rvz)|| p(\rvz)] = \KL[\Dz_G|| \Dz_t] = 0\\ 
% \end{aligned}
% \end{equation}
% Substitute Eq.~\eqref{eq::KL4DA} into Eq.~\eqref{eq::d_{H}_2}, we have:
\begin{equation}
\label{eq::d_{H}_3}
\begin{aligned}
&d_{\gH} (\Dz_s', \Dz_t) \\
&\leq \tau d_{\gH} (\Dz_s,\Dz_t) + (1-\tau) d_{\gH}(\Dz_G, \Dz_t) \\
&\leq \tau d_{\gH} (\Dz_s,\Dz_t) + (1-\tau) d_{\gH}(\Dz_s, \Dz_t) 
= d_{\gH} (\Dz_s,\Dz_t)
\end{aligned}
\end{equation}

\noindent Clearly, if Assumption 4.1 holds with $m > \frac{1}{2}d$, we can prove that $J(m',d) < j(m,d)$ and $d_{\gH}(\Dz'_{s}, \Dz_t) < d_{\gH}(\Dz_{s}, \Dz_t)$. 
Furthermore, it can easily be seen that $\epsilon_{\gS^{'}}(h)) < \epsilon_{\gS}(h))$ and $\underset{h}{\min} \left( \epsilon_{\gT}(h) + \epsilon_{\gS'}(h) \right ) < \underset{h}{\min} \left( \epsilon_{\gT}(h) + \epsilon_{\gS}(h) \right )$ if $(\epsilon_{\gS}(h) - \epsilon_{\gS^{'}}(h)) > 0$ is bounded.
Combining Eq.~\eqref{eq::generated_D}, it is now obvious that:
    \begin{equation}
    \begin{aligned} 
    % \epsilon_{\gT}(h) 
    % &\leq \varepsilon_{\gS'}(h) 
    % + \sqrt{\frac{4}{m'} \left( d \log\frac{2 e m' }{d} + \log \frac{4}{\delta} \right) }  
    %  %
    %  + d_{\gH} (\Dz_{s}^{'}, \Dz_{t}) + \lambda' \\
    %  & \leq \epsilon_{\gS}(h)  + \sqrt{\frac{4}{m} \left( d \log\frac{2 e m }{d} + \log \frac{4}{\delta} \right) } + d_{\gH} (\Dz_{S}, \Dz_{T}) + \lambda
    %%%%%%%%%%%%%%%%%%%%%%%%%%%%%%%%%%%%%%%%%%%%%%%%%%%%%%%%%%%%%%%%%%%%%
\epsilon_{\gT}(h) 
    &\leq \hat{\epsilon}_{\gS'}(h) 
    + J(m',d)
     + d_{\gH} (\Dz_{s}^{'}, \Dz_{t}) + \lambda' \\
    & \leq \hat{\epsilon}_{\gS}(h)  
     + J(m,d)
     + d_{\gH} (\Dz_{s}, \Dz_{t}) + \lambda
    \end{aligned}
    \end{equation}
This completes the proof.
\end{proof}
\end{document}